\documentclass[lettersize,journal]{IEEEtran}
\usepackage{amsmath,amsfonts}
\usepackage{algorithm}
\usepackage{array}
\usepackage{textcomp}
\usepackage{stfloats}
\usepackage{url}
\usepackage{verbatim}
\usepackage{cite}
\usepackage{caption}
\usepackage{graphicx}
\usepackage{subcaption}
\usepackage{adjustbox}
\usepackage{afterpage}
\usepackage{times}
\usepackage{soul}
\usepackage{url}
\usepackage[hidelinks]{hyperref}
\usepackage[utf8]{inputenc}
\usepackage{amsthm}
\usepackage{algpseudocode}
\usepackage{multirow}%
\usepackage{mathrsfs}%
\usepackage{xcolor}%
\usepackage{textcomp}%
\usepackage{manyfoot}%
\usepackage{booktabs}%
\usepackage{listings}%
\usepackage{hhline}
\usepackage{soul}

\begin{document}

\title{DACAD: Domain Adaptation Contrastive Learning for
Anomaly Detection in Multivariate Time Series}

\author{\IEEEauthorblockN{Zahra Zamanzadeh Darban\IEEEauthorrefmark{1}, Yiyuan Yang\IEEEauthorrefmark{2}, Geoffrey~I.~Webb\IEEEauthorrefmark{3}, Charu C.~Aggarwal\IEEEauthorrefmark{4}, Qingsong Wen\IEEEauthorrefmark{5}, Shirui Pan\IEEEauthorrefmark{6}, and Mahsa Salehi\IEEEauthorrefmark{3}}
\thanks{\IEEEauthorrefmark{1}Zahra Zamanzadeh Darban is with the Department of Data Science and AI, Monash University, Melbourne, Australia (email: zahra.zamanzadeh@monash.edu). She is the corresponding author.}
\thanks{\IEEEauthorrefmark{2}Yiyuan Yang is with the Department of Computer Science, University of Oxford, OX1 3SA, Oxford, UK. (email: yiyuan.yang@cs.ox.ac.uk)}
\thanks{\IEEEauthorrefmark{3}Geoffrey~I.~Webb is with the Department of Data Science and AI, Monash University, Australia (email: geoff.webb@monash.edu).}
\thanks{\IEEEauthorrefmark{4}Charu C.~Aggarwal is with the IBM T. J. Watson Research Center, Yorktown Heights, USA (email: charu@us.ibm.com).}
\thanks{\IEEEauthorrefmark{5}Qingsong Wen is with Squirrel Ai Learning, Bellevue, USA. (email: qingsongedu@gmail.com)}
\thanks{\IEEEauthorrefmark{6}Shirui Pan is with the School of ICT, Griffith University, Australia (email: s.pan@griffith.edu.au)}
\thanks{\IEEEauthorrefmark{3}Mahsa Salehi is with the Department of Data Science and AI, Monash University, Australia (email: mahsa.salehi@monash.edu).}
}

\markboth{IEEE Transactions on on Knowledge and Data Engineering,~Vol.~XX, No.~Y, Month~Year}%
{Darban \MakeLowercase{\textit{et al.}}: DACAD: Domain Adaptation Contrastive Learning for Anomaly Detection in Multivariate Time Series}

\maketitle

\begin{abstract}
In time series anomaly detection (TSAD), the scarcity of labeled data poses a challenge to the development of accurate models. Unsupervised domain adaptation (UDA) offers a solution by leveraging labeled data from a related domain to detect anomalies in an unlabeled target domain. However, existing UDA methods assume consistent anomalous classes across domains. To address this limitation, we propose a novel Domain Adaptation Contrastive learning model for Anomaly Detection in multivariate time series (DACAD), combining UDA with contrastive learning. DACAD utilizes an anomaly injection mechanism that enhances generalization across unseen anomalous classes, improving adaptability and robustness. Additionally, our model employs supervised contrastive loss for the source domain and self-supervised contrastive triplet loss for the target domain, ensuring comprehensive feature representation learning and domain-invariant feature extraction. Finally, an effective Center-based Entropy Classifier (CEC) accurately learns normal boundaries in the source domain. Extensive evaluations on multiple real-world datasets and a synthetic dataset highlight DACAD's superior performance in transferring knowledge across domains and mitigating the challenge of limited labeled data in TSAD. 
\end{abstract}


\section{Introduction}
Unsupervised Domain Adaptation (UDA) is a technique used to transfer knowledge from a labeled source domain to an unlabeled target domain. Such an approach is particularly useful when labeled data in the target domain is scarce or unavailable~\cite{xie2022collaborative}. 
Deep learning methods have become the predominant approach in UDA, offering advanced capabilities and significantly improved performance compared to traditional techniques. UDA is crucial in situations where the performance of deep models drops significantly due to the discrepancy between the data distributions in the source and target domains, a phenomenon known as domain shift~\cite{liu2022deep}.

Applications of UDA are diverse, ranging from image and video analysis to natural language processing and time series data analysis. However, the time series domain differs greatly from the image and text data domains for which UDA methods are well developed~\cite{ismail2019deep,wen2020time}. UDA can be particularly challenging for time series analysis, given that often i) the aspects of time series that may be relevant to different tasks can vary greatly, such as frequency, magnitude, rate of change in either of these, global shape or local shape~\cite{wilson2020survey,liu2022deep,xu2024calibrated} and ii) the number of anomalous classes changes between the source and target domains.

For time series data specifically, UDA methods often employ neural architectures as feature extractors. These models are designed to handle domain adaptation, primarily target time series regression and classification problems~\cite{purushotham2016variational,cai2021time,ozyurt2023contrastive,wilson2020multi}, focusing on aligning the major distributions of two domains. This approach may lead to negative transfer effects \cite{zhang2022survey} on minority distributions, which is a critical concern in time series anomaly detection (TSAD). The minority distributions, often representing rare events or anomalies, may be overshadowed or incorrectly adapted due to the model's focus on aligning dominant data distributions, potentially leading to a higher rate of false negatives in anomaly detection~\cite{zhou2024label}.
Since transferring knowledge of anomalies requires aligning minority distributions, anomaly label spaces often have limited similarity across domains, and existing methods face limitations in addressing anomaly detection in time series data. This highlights a significant gap in current UDA methodologies, pointing towards the need for novel approaches that can effectively address the unique requirements of anomaly detection in time series data.

To further substantiate our approach, theoretical insights indicate that reducing contrastive loss lowers the Class-wise Mean Maximum Discrepancy (CMMD)~\cite{quintana2025bridging}, thereby improving domain adaptation. This reinforces the validity of integrating contrastive learning with UDA in our approach and strengthens the theoretical foundation of our work.

Furthermore, in the realm of time series anomaly detection, in the recent model called ContextDA~\cite{lai2023context}, the discriminator aligns source/target domain windows without leveraging label information in the source domain, which makes the alignment ineffective. Unlike these approaches, our proposed model leverages source labels and anomaly injection to enhance feature extraction, particularly by focusing on distinguishing normal from anomalous samples through contrastive learning. This anomaly injection follows the method that has been proposed in the CARLA model~\cite{darban2023carla}, demonstrating its efficacy in real-world scenarios.
Our model leverages source labels and anomaly injection for better feature extraction, enhancing the alignment of normal samples. This is particularly vital due to the challenges in aligning anomalies across domains with different anomalous classes.

In our study, we introduce the Domain Adaptation Contrastive learning model for Anomaly Detection in time series (DACAD), a unique framework for UDA in multivariate time series (MTS) anomaly detection, which leverages contrastive learning (CL) and focuses on contextual representations. It forms positive pairs based on proximity and negative pairs using anomaly injection, subsequently learning their embeddings with both supervised CL in the source domain and self-supervised CL in the target domain.
To sum up, we make the following contributions:
\begin{itemize}
\item We introduce DACAD, a pioneering contrastive learning framework for multivariate time series anomaly detection with UDA. DACAD is a novel deep UDA model that incorporates both supervised contrastive learning in the source domain and self-supervised contrastive learning in the target domain. To address the scarcity of real anomalies, we enhance representation learning through synthetic anomaly injection, providing effective differentiation between normal and anomalous patterns. 
The source code of DACAD is available on GitHub\footnote{\url{https://github.com/zamanzadeh/DACAD}}.
\item Our proposed Center-based Entropy Classifier (CEC) introduces a novel enhancement to DeepSVDD~\cite{ruff2018deep} by explicitly leveraging source domain label information to refine anomaly detection. CEC enforces spatial separation in the feature space by pulling ``normal'' sample representations closer to the center and distancing anomalous ones. This spatial adjustment is guided by a distance metric that enhances anomaly detection by improving class separability.
\item Our comprehensive evaluation with real-world datasets highlights DACAD's efficiency. In comparison with the recent TSAD deep models and UDA models for time series classification and UDA for anomaly detection, DACAD demonstrates superior performance, emphasizing the effective contribution of our approach.
\end{itemize}

\section{Related Works}\label{related}
In this section, we provide a brief survey of the existing literature that intersects with our research. 
We concentrate on two areas: UDA for time series and deep learning for TSAD.

\textbf{UDA for time series:} 
In the realm of UDA for time series analysis, the combination of domain adaptation techniques with the unique properties of time series data presents both challenges and opportunities~\cite{li2023few}. Traditional approaches such as Maximum Mean Discrepancy (MMD)~\cite{long2018conditional} and adversarial learning frameworks~\cite{tzeng2017adversarial} are geared towards minimizing domain discrepancies by developing domain-invariant features. These methods are vital in applications spanning medical, industrial, and speech data, with significant advancements in areas like sleep classification~\cite{zhao2021unsupervised}, energy safety detection~\cite{yang2021pipeline}, arrhythmia classification~\cite{wang2021inter}, and various forms of anomaly detection~\cite{lai2023context}, fault diagnosis~\cite{lu2021new}, and lifetime prediction~\cite{ragab2020contrastive}.

Time series-specific UDA approaches like variational recurrent adversarial deep domain adaptation (VRADA)~\cite{purushotham2016variational}, pioneered UDA for MTS, utilizing adversarial learning with an LSTM network~\cite{hochreiter1997long} and variational RNN~\cite{chung2015recurrent} feature extractor. Convolutional deep domain adaptation for time series (CoDATS)~\cite{wilson2020multi}, built on VRADA's adversarial training, but employed a convolutional neural network as the feature extractor. A metric-based method, time series sparse associative structure alignment (TS-SASA)~\cite{cai2021time} aligns intra and inter-variable attention mechanisms between domains using MMD. Adding to these, the CL for UDA of time series (CLUDA)~\cite{ozyurt2023contrastive} model offers a novel approach, enhancing domain adaptation capabilities in time series. All these methods share a common objective of aligning features across domains, each contributing unique strategies to the complex challenge of domain adaptation in time series classification data. However, they are ineffective when applied to TSAD tasks. Additionally, ContextDA~\cite{lai2023context} is a TSAD model that applies deep reinforcement learning to optimize domain adaptation, framing context sampling as a Markov decision process. However, it is ineffective when the anomaly classes change between the source and the target. 

Recently, Large Language Models (LLMs) and Pretrained Models (PMs) have gained attention for addressing Unsupervised Domain Adaptation (UDA) in time series tasks~\cite{zhou2023one,liu2024large}. These models leverage their robust transfer learning capabilities and the benefits of extensive pre-training on a vast array of sequential data across different domains. This pre-training phase equips the models with comprehensive knowledge and advanced pattern recognition skills, allowing them to effectively adapt to and perform well on specific time series tasks through a relatively straightforward fine-tuning process~\cite{zhang2024large,jin2023large}. This extensive pre-training enables LLMs and PMs to adeptly handle time series anomaly detection, even when these tasks differ significantly from the domains on which they were originally trained. Additionally, diffusion model-based UDA methods have emerged as a promising approach, utilizing the inherent strengths of diffusion models in capturing dynamic changes over time~\cite{yang2024survey}. By integrating a small amount of target domain data to guide the adaptation process, these models can achieve effective domain transfer~\cite{wang2024drift,zhao2024diffusion}. However, neither LLM nor diffusion models are without their challenges. Issues such as computational efficiency and training stability are prevalent and continue to be significant areas of research and discussion in the field~\cite{yang2024survey,jin2023large}.

\textbf{Deep Learning for TSAD:} 
The field of Time Series Anomaly Detection (TSAD) has advanced significantly, embracing a variety of methods ranging from basic statistical approaches to sophisticated deep learning techniques~\cite{schmidl2022anomaly,audibert2022deep,xu2023deep}. Notably, deep learning has emerged as a promising approach due to its autonomous feature extraction capabilities~\cite{darban2022deep}. TSAD primarily focuses on unsupervised~\cite{yue2022ts2vec,hundman2018detecting,audibert2020usad,xu2022anomaly} and semi-supervised methods~\cite{niu2020lstm,zhan2024hfn,zhou2023semi} to tackle the challenge of limited labeled data availability. Unsupervised methods like OmniAnomaly~\cite{su2019robust} and GDN~\cite{deng2021graph} are especially valuable in scenarios with sparse anomaly labels, whereas semi-supervised methods leverage available labels effectively.

Advanced models such as LSTM-NDT~\cite{hundman2018detecting} and THOC~\cite{shen2020timeseries} excel in minimizing forecasting errors and capturing temporal dependencies. However, models like USAD~\cite{audibert2020usad} face challenges with long time series data due to error accumulation in decoding. Additionally, unsupervised representation learning, exemplified by SPIRAL~\cite{lei2019} and TST~\cite{zerveas2021}, shows impressive performance, albeit with scalability issues in long time series. Newer models like TNC~\cite{tonekaboni2021} aim to overcome these challenges using methods such as time-based negative sampling.

Furthermore, contrastive representation learning, crucial in TSAD for pattern recognition, groups similar samples together while distancing dissimilar ones~\cite{zhou2022contrastive}. It has been effectively employed in TS2Vec~\cite{yue2022ts2vec} for multi-level semantic representation and in DCdetector~\cite{yang2023dcdetector}, which uses a dual attention asymmetric design for permutation invariant representations. Recently, PatchAD~\cite{zhong2024patchad} designs a lightweight patch-based MLP-mixer based on contrastive learning for TSAD.


\section{DACAD} \label{sec:dacad}
\textbf{Problem formulation:} Given an unlabeled time series dataset $T$ (target), the problem is to detect anomalies in $T$ using a labeled time series dataset $S$ (source) from a related domain. 

In this section, we present DACAD, which uses temporal correlations ingeniously and adapts to differences between source and target domains. It starts with a labeled dataset in the source domain, featuring both normal and anomalous instances. Both real and synthetic (injected) anomalies aid in domain adaptation, thereby training in a manner that improves generalizability on a wide range of anomaly classes.

\begin{figure*}[t]%
\centering
\includegraphics[width=1.0\textwidth]{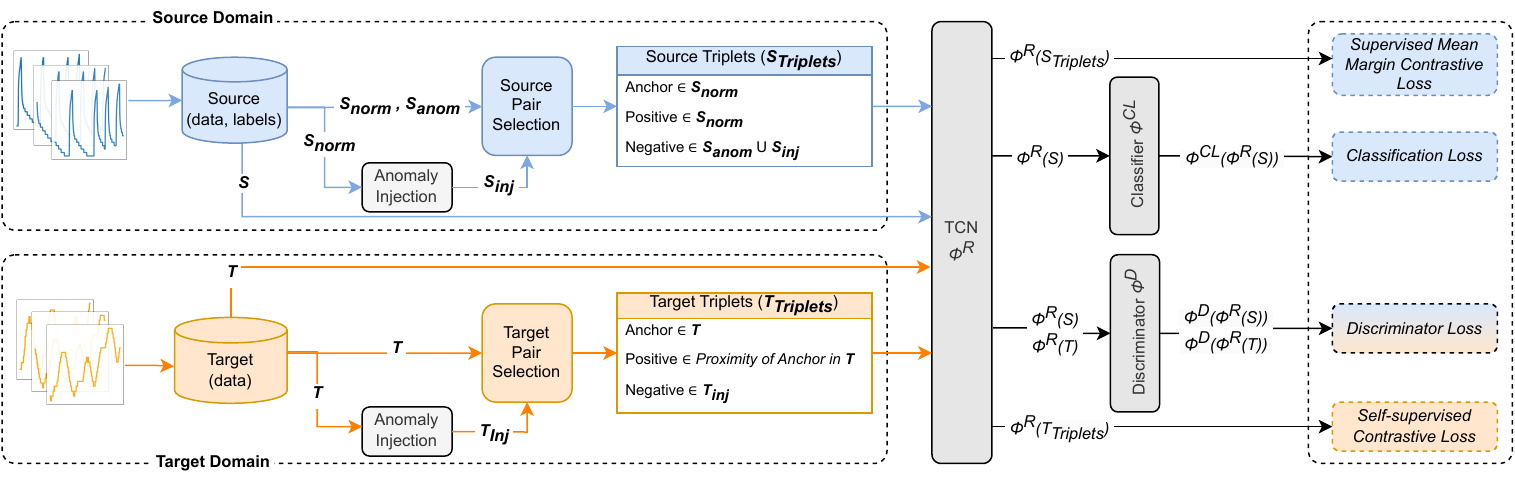}
\caption{DACAD Model Overview: Involves source ($S$) and target ($T$) domains. Source domain uses normal ($S_{\text{norm}}$) and anomalous ($S_{\text{anom}}$) samples, plus synthetic anomalies ($S_{\text{inj}}$) for source triplets ($S_{\text{Triplets}}$) and contrastive loss. Target domain similarly uses proximity-based pair selection and anomaly injection ($T_{\text{inj}}$) to create target triplets ($T_{\text{Triplets}}$). TCN ($\phi^R$) is used for feature extraction. Features from both domains are fed into the discriminator ($\phi^D$) for domain-invariant learning. Source features are classified by classifier $\phi^{CL}$. 
}
\label{fig:arch}
\vspace{-10pt}
\end{figure*}

DACAD's core is CL as inspired by a recent UDA for time series work \cite{ozyurt2023contrastive}, which strengthens its ability to handle changes between domains by improving the feature representation learning in source and target domains. As described in subsection~\ref{sec:rep}, in the target domain, we use a self-supervised contrastive loss~\cite{schroff2015facenet} by forming triplets to minimize the distance between similar samples and maximize the distance between different samples. Additionally, in the source domain, we leverage label information of both anomaly and normal classes and propose an effective supervised contrastive loss~\cite{khosla2020supervised} named supervised mean-margin contrastive loss. A Temporal Convolutional Network (TCN) in DACAD captures temporal dependencies, generating domain-invariant features. A discriminator ensures these features are domain-agnostic, leading to consistent anomaly detection across domains. DACAD's strength lies in its dual ability to adapt to new data distributions and to distinguish between normal and anomalous patterns effectively. In DACAD, time series data is split into overlapping windows of size $W_S$ with a stride of 1, forming detailed source ($S$) and target ($T$) datasets. Source windows are classified as normal ($S_{\text{norm}}$) anomalous $S_{\text{anom}}$) based on anomaly presence. Fig.~\ref{fig:arch} shows the DACAD architecture, and Algorithm \ref{alg:dacad} details its steps. The following subsections explore DACAD's components and their functionalities.
\setlength{\abovecaptionskip}{5pt}

\begin{algorithm}[t]
\footnotesize
\caption{\texttt{DACAD}($S$, $T$, $\alpha$, $\beta$, $\gamma$, $\lambda$)}
\label{alg:dacad}
\begin{algorithmic}[1]
  \renewcommand{\algorithmicrequire}{\textbf{Input:}}
  \renewcommand{\algorithmicensure}{\textbf{Output:}}
    \Require Source time series windows $S = \{w^s_1, w^s_2, ..., w^s_{|S|}\}$, Target time series windows $T = \{w^t_1, w^t_2, ..., w^t_{|T|}$, Loss coefficients $\alpha$, $\beta$ $\gamma$, $\lambda$
    \Ensure Representation function $\phi^R$, classifier $\phi^{CL}$, centre $c$
    
    \State Initialise $\phi^R$, $\phi^D$, $\phi^{CL}$, $c$
    \State Split $S$ to $S_{\text{norm}}$ and $S_{\text{anom}}$
    
    \State Create $S_{\text{inj}}$, $T_{\text{inj}}$ \hfill $\triangleright$ \text{Anomaly Injection (\ref{sec:anomaly_injection})}
    
    \State Form $S_{\text{triplets}}$ and $T_{\text{triplets}}$ \hfill $\triangleright$ \text{Pair Selection (\ref{sec:pair_selection})}

    \For{each training iteration}
        
        \State Compute $\phi^R(S)$,$\phi^R(T)$,$\phi^R(S_{\text{Triplets}})$,$\phi^R(T_{\text{Triplets}})$ \hfill $\triangleright$ \text{(\ref{sec:rep})}
        
        \State Compute $\mathcal{L}_{\text{SupCont}}$ using Eq. (\ref{eq:loss_sup}) and $\phi^R(S_{\text{Triplets}})$
        \State Compute $\mathcal{L}_{\text{SelfCont}}$ using Eq. (\ref{eq:loss_self}) and $\phi^R(T_{\text{Triplets}})$
        \State Compute $\mathcal{L}_{\text{Disc}}$ using Eq. (\ref{eq:loss_disc}), $\phi^R(S)$ and $\phi^R(T)$ \hfill $\triangleright$ \text{(\ref{sec:disc})}
        \State Compute $\mathcal{L}_{\text{Cls}}$ using Eq. (\ref{eq:loss_class}) and $\phi^R(S)$ \hfill $\triangleright$ \text{(\ref{sec:classifier})}
        
        \State 
        \parbox[t]{\dimexpr\linewidth-\algorithmicindent}{%
        $\mathcal{L}_{\text{DACAD}} \gets \alpha \cdot \mathcal{L}_{\text{SupCont}} + \beta \cdot \mathcal{L}_{\text{SelfCont}} + \gamma \cdot \mathcal{L}_{\text{Disc}} + \lambda \cdot \mathcal{L}_{\text{Cls}}$
        }
        
        \State Update model parameters to minimise $\mathcal{L}_{\text{DACAD}}$
    \EndFor
    
    \State \textbf{Return} $\phi^R$, $\phi^{CL}$, $c$
\end{algorithmic}
\end{algorithm} 

\subsection{Anomaly Injection} \label{sec:anomaly_injection}
In the anomaly injection phase, we augment the original time series windows through a process of negative augmentation, thereby generating synthetic anomalous time series windows. This step is applied to all windows in the target domain ($T$) and all normal windows in the source domain ($S_{\text{norm}}$). We employ the anomaly injection method outlined in~\cite{darban2023carla}, which encompasses five distinct types of anomalies: Global, Seasonal, Trend, Shapelet, and Contextual. 
This procedure results in the creation of two new sets of time series windows. The first set, $S_{\text{inj}}$, consists of anomaly-injected windows derived from the normal samples in the source domain. The second set, $T_{\text{inj}}$, comprises anomaly-injected windows originating from the target domain data.

\subsection{Pair Selection} \label{sec:pair_selection}
In DACAD's pair selection step, appropriate triplets from the source and target domains are created for CL. In the source domain, we use labels to form distinct lists of normal samples ($S_{\text{norm}}$), anomalous samples ($S_{\text{anom}}$), and anomaly-injected samples ($S_{\text{inj}}$). This allows for a supervised CL approach, enhancing differentiation between normal and anomalous samples. Here, triplets ($S_{\text{triplets}}$) consist of an anchor (normal window from $S_{\text{norm}}$), a positive (different normal window from $S_{\text{norm}}$), and a negative (randomly selected from either an anomalous window from $S_{\text{anom}}$ or an anomaly-injected anchor from $S_{\text{inj}}$).

In the unlabeled target domain, a self-supervised approach constructs the triplets ($T_{\text{triplets}}$). Each target triplet includes an anchor (original window from $T$), a positive (temporally close window to anchor, from $T$, likely sharing similar characteristics), and a negative (anomaly-injected anchor from $T_{\text{inj}}$).

\subsection{\texorpdfstring{Representation Layer ($\phi^R$)}{Representation Layer (phi^R)}} \label{sec:rep}

In our model, we employ a TCN~\cite{lea2016temporal} for the representation layer, which is adept at handling MTS windows. This choice is motivated by the TCN's ability to capture temporal dependencies effectively, a critical aspect in time series analysis. The inputs to the TCN are the datasets and triplets from both domains, specifically $S$, $T$, $S_{\text{Triplets}}$, and $T_{\text{Triplets}}$. The outputs of the TCN, representing the transformed feature space, are 
\begin{itemize}
    \item $\phi^R(S)$: The representation of source windows.
    \item $\phi^R(T)$: The representation of target windows.
    \item $\phi^R(S_{\text{Triplets}})$: The representation of source triplets.
    \item $\phi^R(T_{\text{Triplets}})$: The representation of target triplets.
\end{itemize}

Utilizing the outputs from the representation layer $\phi^R$, we compute two distinct loss functions. These are the supervised mean margin contrastive loss for source domain data ($\mathcal{L}_{\text{SupCont}}$) and the self-supervised contrastive triplet loss for target domain ($\mathcal{L}_{\text{SelfCont}}$). These loss functions are critical for training our model to differentiate between normal and anomalous patterns in both source and target domains.

\subsubsection{Supervised Mean Margin Contrastive Loss for Source Domain ($\mathcal{L}_{\text{SupCont}}$)}
This loss aims to embed time series windows into a feature space where normal sequences are distinctively separated from anomalous ones. It utilizes a triplet loss framework, comparing a base (anchor) window with both normal (positive) and anomalous (negative) windows~\cite{khosla2020supervised}. Our method diverges from traditional triplet loss by focusing on the average effect of all negatives within a batch. The formula is given by Equation~(\ref{eq:loss_sup}):

\begin{footnotesize}
\begin{equation}
\label{eq:loss_sup}
\begin{aligned}
    &\mathcal{L}_{\text{SupCont}} = \frac{1}{|B|} \sum_{i=1}^{|B|} \max \\
    &\left(\frac{1}{|N|}\sum_{j=1}^{|N|} \left(\| \phi^R(a^s_i) - \phi^R(p^s_i) \|_2^2
    - \| \phi^R(a^s_i) - \phi^R(n^s_j) \|_2^2 + m\right), 0 \right) \\
\end{aligned}
\end{equation}   
\end{footnotesize}
Here, $|B|$ is the batch size, and $|N|$ is the number of negative samples in the batch. The anchor $a^s_i$ is a normal time series window, and the positive pair $p^s_i$ is another randomly selected normal window. Negative pairs $n^s_j$ are either true anomalous windows or synthetically created anomalies through the anomaly injector module applied on the anchor $a^s_i$. This loss function uses true anomaly labels to enhance the separation between normal and anomalous behaviors by at least the margin $m$. It includes both genuine and injected anomalies as negatives, balancing ground truth and potential anomaly variations. The supervised mean margin contrastive loss offers several advantages for TSAD: supervised learning (uses labels for better anomaly distinction), comprehensive margin (applies a margin over average distance to negatives), and flexible negative sampling (incorporates a mix of real and injected anomalies, enhancing robustness against diverse anomalous patterns).

\subsubsection{Self-supervised Contrastive Triplet Loss for Target Domain ($\mathcal{L}_{\text{SelfCont}}$)} \label{sec:selfcont}
For the target domain, we employ a self-supervised contrastive approach using triplet loss~\cite{schroff2015facenet}, designed to ensure that the anchor window is closer to a positive window than to a negative window by a specified margin. The anchor is assumed to be normal, following the standard TSAD assumption that the majority of the data is normal, as anomalies are rare by nature. The self-supervised triplet loss formula is shown in Equation (\ref{eq:loss_self}):

\begin{footnotesize}
\begin{equation}
\label{eq:loss_self}
\begin{aligned}
    &\mathcal{L}_{\text{SelfCont}} = \\
    &\frac{1}{|B|} \sum_{i=1}^{|B|} \max \left(\| \phi^R(a^t_i) - \phi^R(p^t_i) \|_2^2 - \| \phi^R(a^t_i) - \phi^R(n^t_i) \|_2^2 + m, 0 \right)
\end{aligned}
\end{equation}
\end{footnotesize}
In this setup, the anchor window $a^t_i$ is compared to a positive window $p^t_i$ (a randomly selected nearby window from $T$) and a negative window $n^t_i$ (the anomaly-injected version of the anchor from $T_{\text{inj}}$), ensuring the anchor is closer to the positive than the negative by at least the margin $m$.

\subsection{\texorpdfstring{Discriminator Component ($\phi^D$)}{Discriminator Component (phi^D)}} \label{sec:disc}
The model incorporates a discriminator component within an adversarial framework. This component is crucial for ensuring that the learned features are not only relevant for the anomaly detection task but also general enough to be applicable across different domains. The discriminator is specifically trained to distinguish between representations from the source and target domains.

Designed to differentiate between features extracted from the source and target domains, the discriminator employs adversarial training techniques similar to those found in Generative Adversarial Networks~\cite{creswell2018generative}. Discriminator $\phi^D$ is trained to differentiate between source and target domain features, while $\phi^R$ is conditioned to produce domain-invariant features.

A crucial element is the Gradient Reversal Layer~\cite{ganin2015unsupervised}, which functions normally during forward passes but reverses gradient signs during backpropagation. This setup enhances the training of $\phi^D$ and simultaneously adjusts $\phi^R$ to produce features that challenge $\phi^D$.

The discriminator's training involves balancing its loss against other model losses. Effective domain adaptation occurs when $\phi^R$ yields discriminator accuracy close to random guessing. $\phi^D$, taking $\phi^R(S)$ and $\phi^R(T)$ as inputs, classifies them as belonging to the source or target domain. Its loss, a binary classification problem, minimizes classification error for source and target representations. The loss function for the discriminator $\phi^D$ is defined using the Binary Cross-Entropy (BCE) loss, as shown in Equation (\ref{eq:loss_disc}):

\begin{footnotesize}
\begin{equation}
\label{eq:loss_disc}
\begin{aligned}    
   &\mathcal{L}_{\text{Disc}} = -\frac{1}{|S| + |T|} \left( \sum_{i=1}^{|S|} \log(f(w_i^s))) + \sum_{j=1}^{|T|} \log(1 - f(w_j^t))) \right) \\
   &where ~f(w) = \phi^{D}(\phi^{R}(w))
\end{aligned}
\end{equation}
\end{footnotesize}
Here, $|S|$ and $|T|$ are the source and target window counts. $w_i^s$ and $w_j^t$ represent the $i^{th}$ and $j^{th}$ windows from $S$ and $T$. The function $\phi^D(\phi^R(\cdot))$ computes the likelihood of a window being from the source domain.

\subsection{\texorpdfstring{Centre-based Entropy Classifier ($\phi^{CL}$)}{}} \label{sec:classifier}
Extending the DeepSVDD~\cite{ruff2018deep} designed for anomaly detection, the CEC ($\phi^{CL}$) in DACAD is proposed as an effective anomaly detection classifier in the source domain. It assigns anomaly scores to time series windows, using labeled data from the source domain $S$ for training and applying the classifier to target domain data $T$ during inference. It is centered around a Multi-Layer Perceptron (MLP) with a unique ``center'' parameter crucial for classification.

The classifier operates by spatially separating transformed time series window representations ($\phi^R(w_i^s)$) in the feature space relative to a learnable ``center'' $c$. The MLP aims to draw normal sample representations closer to $c$ and push anomalous ones further away. This spatial reconfiguration is quantified using a distance metric, forming the basis for anomaly scoring. A sample closer to $c$ is considered more normal, and vice versa. The effectiveness of this spatial adjustment is measured using a BCE-based loss function, expressed in Equation (\ref{eq:loss_class}):

\begin{footnotesize}
\begin{equation}
\label{eq:loss_class}
\begin{aligned}
&\mathcal{L}_{\text{Cls}} =\\
&-\frac{1}{|S|} \sum_{i=1}^{|S|} \left[ y_i \cdot \log(\| g(w_i^s) - c \|_2^2) + (1 - y_i) \cdot \log(1 - \| g(w_i^s) - c \|_2^2) \right] \\
                           & where ~g(w) = \phi^{CL}(\phi^{R}(w))\\
\end{aligned}
\end{equation}
\end{footnotesize}
In this Equation, $|S|$ is the number of samples in $S$, $w_i^s$ is the $i^{th}$ window in $S$, and $y_i$ is its ground truth label, with 1 for normal and 0 for anomalous samples.

The loss function is designed to minimize the distance between normal samples and $c$ while maximizing it for anomalous samples. These distances are directly used as anomaly scores, offering a clear method for anomaly detection.

\subsection{Overall Loss in DACAD} \label{sec:dacad_loss}
The overall loss function in the DACAD model is the amalgamation of four distinct loss components, each contributing to the model's learning process in different aspects. These components are the \textbf{Supervised Contrastive Loss} ($\mathcal{L}_{\text{SupCont}}$), the \textbf{Self-Supervised Contrastive Loss} ($\mathcal{L}_{\text{SelfCont}}$), the \textbf{Discriminator Loss} ($\mathcal{L}_{\text{Disc}}$), and the \textbf{Classifier Loss} ($\mathcal{L}_{\text{Cls}}$). The overall loss function for DACAD denoted as $\mathcal{L}_{\text{DACAD}}$, is formulated as a weighted sum of these components (with a specific weight: $\alpha$ for $\mathcal{L}_{\text{SupCont}}$, $\beta$ for $\mathcal{L}_{\text{SelfCont}}$, $\gamma$ for $\mathcal{L}_{\text{Disc}}$, and $\lambda$ for $\mathcal{L}_{\text{Cls}}$), as shown in Equation (\ref{eq:loss_dacad}):

\begin{footnotesize}
\begin{equation}
\label{eq:loss_dacad}
\begin{aligned}
\mathcal{L}_{\text{DACAD}} = \alpha \cdot \mathcal{L}_{\text{SupCont}} + \beta \cdot \mathcal{L}_{\text{SelfCont}} + \gamma \cdot \mathcal{L}_{\text{Disc}} + \lambda \cdot \mathcal{L}_{\text{Cls}}
\end{aligned}
\end{equation}
\end{footnotesize}

The overall loss function $\mathcal{L}_{\text{DACAD}}$ is what the model seeks to optimize during the training process. By fine-tuning these weights ($\alpha$, $\beta$, $\gamma$, and $\lambda$), the model can effectively balance the significance of each loss component. This balance is crucial as it allows the model to cater to specific task requirements.

\subsection{Role of Anomaly Injection in Convergence}
Under standard assumptions in nonconvex optimization, the anomaly injection mechanism in DACAD enforces a minimum margin $m$ between normal and anomalous embeddings. This guarantees that the contrastive losses remain strictly positive unless the model achieves the desired separation, thereby preventing degenerate solutions such as feature collapse. Assuming that the feature extractor is $L$-Lipschitz and that the loss functions are smooth with Lipschitz-continuous gradients~\cite{bottou2018optimization}, an update via stochastic gradient descent with a sufficiently small learning rate ensures that the overall loss is lower bounded. Moreover, under these conditions, the loss locally satisfies a Polyak–Łojasiewicz condition, which implies that the expected gradient norm decreases at a rate of $\mathcal{O}(1/K)$ over $K$ iterations~\cite{ghadimi2013stochastic}. This result formally guarantees convergence to a stationary point.

\subsection{DACAD’s Inference} \label{sec:dacad_inf}
In the inference phase of the DACAD model, the primary objective is to identify anomalies in the target domain $T$. This is accomplished for each window $w^t$ in $T$. The anomaly detection is based on the concept of spatial separation in the feature space, as established during the training phase. The anomaly score for each window is derived from the spatial distance between the classifier's output and a learnable center $c$ in the feature space. This score is calculated as the squared Euclidean distance, as shown in Equation ~(\ref{eq:ascore}):

\begin{footnotesize}
\begin{equation}
\label{eq:ascore}
\begin{aligned}
{\text{Anomaly Score}}(w^t) = \| \phi^{CL}(\phi^{R}(w^t)) - c \|_2^2
\end{aligned}
\end{equation}
\end{footnotesize}
Where $\phi^{CL}(\phi^{R}(w^t))$ denotes the feature representation of the window $w^t$ after processing by the representation layer $\phi^R$ and the classifier layer $\phi^{CL}$. The distance measured here quantifies how much each window in the target domain deviates from the ``normal'' pattern, as defined by the center $c$. 

The anomaly score is a crucial metric in this process. A higher score suggests a significant deviation from the normative pattern, indicating a higher likelihood of the window being an anomaly. Conversely, a lower score implies that the window's representation is closer to the center, suggesting it is more likely to be normal. 
In practical terms, the anomaly score can be thresholded to classify windows as either normal or anomalous. By effectively utilizing these anomaly scores, the DACAD model provides a robust mechanism for identifying anomalous patterns in unlabeled target domain data.

\section{Experiments}\label{sec:experiments}
This section provides a comprehensive evaluation of DACAD, covering the experimental setups (Section \ref{sec:setup}) and results (Section \ref{sec:compare} and \ref{sec:ablation}) to clearly understand its performance and capabilities in different contexts.

\begin{table}[t]
\renewcommand{\arraystretch}{1.2}
\centering
\caption{Statistics of the benchmark datasets used.}
\label{tab:dss}
\resizebox{\linewidth}{!}{%
\begin{tabular}{c|ccccc}
\hline
\textbf{Benchmark} & \textbf{\# datasets} & \textbf{\# dims} & \textbf{Train size} & \textbf{Test size} & \textbf{Anomaly ratio} \\ 
\hline
MSL~\cite{hundman2018detecting}         & 27 & 55 & 58,317  & 73,729  & 10.72\% \\
SMAP~\cite{hundman2018detecting}       & 55 & 25 & 140,825 & 444,035 & 13.13\% \\
SMD~\cite{su2019robust}         & 28 & 38 & 708,405 & 708,420 & 4.16\%  \\
Boiler~\cite{cai2021time}         & 3 & 274 & 277,152 & 277,152 & 15.00\%  \\
\hline
\end{tabular}
}%
\vspace{-10pt}
\end{table}

\subsection{Datasets} \label{datasets}
We evaluate the proposed model and make comparisons across the four datasets, including the three most commonly used real benchmark datasets for TSAD and the dataset used for time series domain adaptation. The datasets are summarized in Table~\ref{tab:dss}. Since DACAD utilizes UDA, it is designed to work in scenarios where datasets contain multiple entities. Domain adaptation relies on inter-domain feature alignment, requiring at least one entity as the source and one as the target.

\textbf{Mars Science Laboratory (MSL)} and \textbf{Soil Moisture Active Passive (SMAP)}\footnote{\url{https://www.kaggle.com/datasets/patrickfleith/nasa-anomaly-detection-dataset-smap-msl}} datasets~\cite{hundman2018detecting} are real-world datasets collected from NASA spacecraft. These datasets contain anomaly information derived from reports of incident anomalies for a spacecraft monitoring system. MSL and SMAP comprise 27 and 55 datasets, respectively, and each is equipped with a predefined train/test split, where, unlike other datasets, their training set is unlabeled.

\textbf{Server Machine Dataset (SMD)}\footnote{\url{https://github.com/NetManAIOps/OmniAnomaly/tree/master/ServerMachineDataset}}~\cite{su2019robust} is gathered from 28 servers, incorporating 38 sensors, over a span of 10 days. During this period, normal data was observed within the initial 5 days, while anomalies were sporadically injected during the subsequent 5 days. The dataset is also equipped with a predefined train/test split, where the training data is unlabeled.

\textbf{Boiler Fault Detection Dataset.}\footnote{\url{https://github.com/DMIRLAB-Group/SASA-pytorch/tree/main/datasets/Boiler}}~\cite{cai2021time} The Boiler dataset includes sensor information from three separate boilers, with each boiler representing an individual domain. The objective of the learning process is to identify the malfunctioning blowdown valve in each boiler. Obtaining samples of faults is challenging due to their scarcity in the mechanical system. 

\section{Baselines} \label{baselines}
Below, we will provide an enhanced description of the UDA models for time series classification and anomaly detection.
Additionally, we provide a description of the five prominent and state-of-the-art TSAD models that were used for comparison with DACAD. We have selected the models from different categories of TSAD, namely, unsupervised reconstruction-based (OmniAnomaly~\cite{su2019robust} and AnomalyTransformer~\cite{xu2022anomaly}) models, unsupervised forecasting-based (THOC~\cite{shen2020timeseries}), and self-supervised contrastive learning (TS2Vec~\cite{yue2022ts2vec} and DCdetector~\cite{yang2023dcdetector}) TSAD models.

\textbf{AE-MLP}~\cite{sakurada2014anomaly} is an anomaly detection model that uses autoencoders for nonlinear dimensionality reduction. It encodes data points into a lower-dimensional space and reconstructs them, highlighting deviations to detect anomalies.

\textbf{AE-LSTM}~\cite{malhotra2016lstm} is an anomaly detection model using an LSTM-based encoder-decoder for multi-sensor data. It detects anomalies by the reconstruction error of time series sequences.

\textbf{RDC}\footnote{\url{https://github.com/syorami/DDC-transfer-learning}}~\cite{tzeng2014deep} is built upon deep domain confusion to enhance domain adaptation by maximizing domain invariance. It aligns feature distributions through a domain confusion objective, making learned representations indistinguishable across domains and improving model performance on unseen domains.

\textbf{VRADA}\footnote{\url{https://github.com/floft/vrada}}~\cite{purushotham2016variational} combines deep domain confusion~\cite{tzeng2014deep} with variational recurrent adversarial deep domain adaptation~\cite{purushotham2016variational}, optimizing source domain label prediction, MMD, and domain discrimination using latent representations from the LSTM encoder. Concurrently, AE-LSTM's reconstruction objective is utilized for anomaly detection.

\textbf{SASA}\footnote{\url{https://github.com/DMIRLAB-Group/SASA}}~\cite{cai2021time} improves time series classification using adaptation across domains by identifying and aligning key sparse patterns. It uses a self-attention layer with LSTM units to find optimal global context windows for source and target domains and aligns them using Maximum Mean Discrepancy (MMD).

\textbf{CLUDA}\footnote{\url{https://github.com/oezyurty/CLUDA}}~\cite{ozyurt2023contrastive}
is a novel UDA framework for time series data, leveraging contrastive learning to capture domain-invariant semantics while preserving label information. It is the first UDA framework designed for contextual representation learning in MTS, with effectiveness demonstrated on multiple time series classification datasets.

\textbf{ContextDA}\footnote{There is no implementation available for this model, so we rely on the results claimed in the paper.}~\cite{lai2023context} is a sophisticated approach for detecting anomalies in time series data. It formulates context sampling as a Markov decision process and employs deep reinforcement learning to optimize the domain adaptation process. This model is designed to generate domain-invariant features for better anomaly detection across various domains. It has shown promise in transferring knowledge between similar or entirely different domains.

\textbf{OmniAnomaly}\footnote{\url{https://github.com/smallcowbaby/OmniAnomaly}}~\cite{su2019robust} is a model operating on an unsupervised basis, employing a Variational Autoencoder (VAE) to handle MTS data. It identifies anomalies by evaluating the reconstruction likelihood of specific data windows.

\textbf{THOC}\footnote{We utilized the authors' shared implementation, as it is not publicly available.}~\cite{shen2020timeseries} utilizes a multi-layer dilated recurrent neural network (RNN) alongside skip connections in order to handle contextual information effectively. It adopts a temporal hierarchical one-class network approach for detecting anomalies.

\textbf{TS2Vec}\footnote{\url{https://github.com/yuezhihan/ts2vec}}~\cite{yue2022ts2vec} is an unsupervised model that is capable of learning multiple contextual representations of MTS and UTS semantically at various levels. This model employs contrastive learning in a hierarchical way, which provides a contextual representation. A method within TS2Vec has been proposed for application in TSAD.

\textbf{AnomalyTransformer}\footnote{\url{https://github.com/thuml/Anomaly-Transformer}}~\cite{xu2022anomaly} detects anomalies in time series data by using a novel ``Anomaly-Attention'' mechanism to compute association discrepancies, enhancing the distinguishability between adjacent normal and abnormal points through a minimax strategy.

\textbf{DCdetector}\footnote{\url{https://github.com/DAMO-DI-ML/KDD2023-DCdetector}}~\cite{yang2023dcdetector} is distinctive for its use of a dual attention asymmetric design combined with contrastive learning. Unlike traditional models, DCdetector does not rely on reconstruction loss for training. Instead, it utilizes pure contrastive loss to guide the learning process. This approach enables the model to learn a permutation invariant representation of time series anomalies, offering superior discrimination abilities compared to other methods.

\subsection{Experimental Setup} \label{sec:setup}
In our study, we evaluate several SOTA TSAD models, including OmniAnomaly~\cite{su2019robust}, TS2Vec~\cite{yue2022ts2vec}, THOC~\cite{shen2020timeseries}, and DCdetector~\cite{yang2023dcdetector} and SOTA UDA models that support MTS classification including VRADA~\cite{purushotham2016variational} and CLUDA~\cite{ozyurt2023contrastive} on benchmark datasets previously mentioned in section~\ref{tab:dss} using their source code and best hyper-parameters as they stated to ensure a fair evaluation. The hyperparameters used in our implementation are as follows: DACAD consists of a 3-layer TCN architecture with three different channel sizes [128, 256, 512] to capture temporal dependencies. The dimension of the representation is 1024. We use the same hyperparameters across all datasets to evaluate DACAD: window size ($W_S$) = 100, margin $m$ = 1, and we run our model for 20 epochs.

\setlength{\tabcolsep}{3pt}
\begin{table*}[t]
\caption{F1, AUPR, and AUROC results for various models on multivariate time series benchmark datasets (SMD, MSL, SMAP). The most optimal evaluation results are displayed in bold, while the second-best ones are indicated by \underline{underline}.}
\label{tab:base}
\centering
\footnotesize
\resizebox{\textwidth}{!}{%
\renewcommand{\arraystretch}{1.8}
\begin{tabular}{c|c|c|cccc|cccc|cccc}
\hline
\multirow{2}{*}{\rotatebox[origin=c]{90}{\textbf{Type}}} & \multirow{2}{*}{\rotatebox[origin=c]{90}{\textbf{Model}}} & & \multicolumn{4}{c|}{\textbf{SMD}} 
& \multicolumn{4}{c|}{\textbf{MSL}} & \multicolumn{4}{c}{\textbf{SMAP}}     \\ \hhline{~|~|~|------------}
&& \textbf{src} & \textbf{1-1}  & \textbf{2-3}  & \textbf{3-7}  & \textbf{1-5}  & \textbf{F-5}  & \textbf{P-10} & \textbf{D-14} & \textbf{C-1}  & \textbf{A-7}  & \textbf{P-2}  & \textbf{E-8}  & \textbf{D-7}  \\ 
\hline
\multirow{8}{*}{\rotatebox[origin=c]{90}{\textbf{Classification}}} &
\multirow{4}{*}{\rotatebox[origin=c]{90}{\textbf{VRADA}~\cite{purushotham2016variational}}} &
F1                & \underline{0.511} & 0.435 & 0.268 & 0.291 & 0.302 & 0.321 & 0.310 & 0.285 & 0.244 & 0.282 & 0.349 & 0.226 \\ 
&& AUPR              & \underline{0.523$\pm$0.223} & 0.433$\pm$0.172 & 0.316$\pm$0.208 & 0.292$\pm$0.147 & 0.201$\pm$0.133 & 0.199$\pm$0.159 & \underline{0.351$\pm$0.244} & 0.167$\pm$0.132 & 0.133$\pm$0.196 & \underline{0.265$\pm$0.263} & 0.261$\pm$0.247 & 0.120$\pm$0.149 \\
&& AUROC             & \underline{0.802$\pm$0.130} & 0.731$\pm$0.121 & 0.597$\pm$0.146 & 0.651$\pm$0.088 & 0.526$\pm$0.054 & 0.516$\pm$0.097 & 0.615$\pm$0.207 & 0.506$\pm$0.024 & 0.474$\pm$0.176 & \underline{0.573$\pm$0.162} & 0.612$\pm$0.120 & 0.443$\pm$0.170 \\
&& Aff-F1 & \textbf{0.840$\pm$0.128} & \underline{0.844$\pm$0.110} & \underline{0.830$\pm$0.117} & \underline{0.807$\pm$0.123} & 0.697$\pm$0.041 & 0.701$\pm$0.053 & 0.721$\pm$0.072 & 0.714$\pm$0.066 & 0.737$\pm$0.135 & \underline{0.787$\pm$0.148} & \underline{0.780$\pm$0.139} & \underline{0.768$\pm$0.132}\\
\hhline{~|--------------}
&\multirow{4}{*}{\rotatebox[origin=c]{90}{\textbf{CLUDA}~\cite{ozyurt2023contrastive}}} &
F1                & 0.435 & \underline{0.487} & 0.320 & 0.314 & \underline{0.395} & \underline{0.368} & \underline{0.324} & 0.312 & 0.292 & 0.278 & \underline{0.384} & 0.293 \\ 
&& AUPR & 0.423$\pm$0.223 & \underline{0.494$\pm$0.174} & \underline{0.400$\pm$0.201} & \underline{0.328$\pm$0.140} & \underline{0.325$\pm$0.263} & 0.239$\pm$1617 & 0.318$\pm$0.276 & 0.193$\pm$1640 & \underline{0.250$\pm$0.290} & 0.242$\pm$0.217 & \underline{0.332$\pm$0.263} & \underline{0.193$\pm$0.220} \\
&& AUROC             & 0.788$\pm$0.124 & \underline{0.794$\pm$0.121} & \underline{0.725$\pm$0.155} & \underline{0.693$\pm$0.119} & \underline{0.579$\pm$0.204} & 0.569$\pm$0.132 & 0.558$\pm$0.222 & 0.479$\pm$0.164 & \underline{0.560$\pm$0.223} & 0.500$\pm$0.209 & \underline{0.692$\pm$0.171} & 0.480$\pm$0.222 \\
&& Aff-F1 & 0.816$\pm$0.103 & 0.823$\pm$0.105 & 0.813$\pm$0.101 & 0.791$\pm$0.124 & 0.760$\pm$0.090 & 0.743$\pm$0.082 & 0.748$\pm$0.093 & 0.735$\pm$0.088 & 0.738$\pm$0.099 & 0.753$\pm$0.114 & 0.757$\pm$0.123 & 0.757$\pm$0.128 \\
\hline
\multirow{20}{*}{\rotatebox[origin=c]{90}{\textbf{Anomaly Detection}}} &
\multirow{4}{*}{\rotatebox[origin=c]{90}{\textbf{Omni}~\cite{su2019robust}}} &
F1                & 0.397 & 0.413 & \underline{0.419} & \underline{0.421} & 0.254 & 0.266 & 0.246 & 0.242 & 0.320 & 0.326 & 0.331 & 0.318 \\
&& AUPR              & 0.296$\pm$0.155 & 0.309$\pm$0.165 & 0.314$\pm$0.171 & 0.318$\pm$0.182 & 0.154$\pm$0.184 & 0.154$\pm$0.184 & 0.152$\pm$0.185 & 0.148$\pm$0.186 & 0.111$\pm$0.129 & 0.114$\pm$0.130 & 0.116$\pm$0.131 & 0.112$\pm$0.131 \\
&& AUROC             & 0.619$\pm$0.153 & 0.628$\pm$0.164 & 0.639$\pm$0.156 & 0.620$\pm$0.155 & 0.392$\pm$0.173 & 0.392$\pm$0.173 & 0.398$\pm$0.165 & 0.386$\pm$0.175 & 0.421$\pm$0.189 & 0.406$\pm$0.188 & 0.406$\pm$0.190 & 0.414$\pm$0.181 \\
&& Aff-F1 & 0.635$\pm$0.100 & 0.640$\pm$0.097 & 0.639$\pm$0.098 & 0.638$\pm$0.099 & 0.680$\pm$0.046 & 0.682$\pm$0.047 & 0.679$\pm$0.047 & 0.681$\pm$0.049 & 0.700$\pm$0.089 & 0.701$\pm$0.089 & 0.701$\pm$0.089 & 0.700$\pm$0.089 \\
\hhline{~|--------------}
&\multirow{4}{*}{\rotatebox[origin=c]{90}{\textbf{THOC}~\cite{shen2020timeseries}}} &
F1                & 0.156 & 0.167 & 0.168 & 0.168 & 0.307 & 0.317 & 0.309 & 0.310 & 0.321 & 0.328 & 0.335 & 0.320 \\
&& AUPR              & 0.090$\pm$0.095 & 0.106$\pm$0.129 & 0.109$\pm$0.128 & 0.109$\pm$0.128 & 0.241$\pm$0.278 & \underline{0.247$\pm$0.275} & 0.244$\pm$0.277 & \underline{0.242$\pm$2778} & 0.191$\pm$0.263 & 0.197$\pm$0.264 & 0.200$\pm$0.266 & 0.191$\pm$0.264 \\
&& AUROC             & 0.646$\pm$1.550 & 0.646$\pm$0.154 & 0.657$\pm$0.161 & 0.648$\pm$0.158 & 0.631$\pm$0.177 & \underline{0.637$\pm$0.182} & \underline{0.637$\pm$0.182} & \underline{0.640$\pm$0.183} & 0.535$\pm$0.211 & 0.539$\pm$0.211 & 0.543$\pm$0.211 & \underline{0.539$\pm$0.213} \\
&& Aff-F1 & 0.675$\pm$0.125 & 0.675$\pm$0.126 & 0.676$\pm$0.126 & 0.673$\pm$0.125 & 0.777$\pm$0.119 & 0.775$\pm$0.118 & 0.775$\pm$0.118 & 0.778$\pm$0.119 & 0.780$\pm$0.152 & 0.777$\pm$0.149 & 0.778$\pm$0.151 & 0.779$\pm$0.152 \\
\hhline{~|--------------}
& \multirow{4}{*}{\rotatebox[origin=c]{90}{\textbf{TS2Vec}~\cite{yue2022ts2vec}}} &
F1 & 0.171 & 0.173 & 0.173 & 0.173 & 0.320 & 0.316 & 0.317 & \underline{0.313} & \underline{0.362} & \underline{0.368} & 0.374 & \underline{0.362} \\
&& AUPR & 0.112$\pm$0.076 & 0.112$\pm$0.076 & 0.116$\pm$0.075 & 0.116$\pm$0.075 & 0.137$\pm$0.137 & 0.138$\pm$0.136 & 0.135$\pm$0.138 & 0.133$\pm$0.138 & 0.145$\pm$0.167 & 0.147$\pm$0.167 & 0.149$\pm$0.168 & 0.143$\pm$0.165 \\
&& AUROC & 0.492$\pm$0.046 & 0.493$\pm$0.045 & 0.489$\pm$0.044 & 0.487$\pm$0.040 & 0.511$\pm$0.096 & 0.514$\pm$0.096 & 0.514$\pm$0.096 & 0.513$\pm$0.096 & 0.504$\pm$0.088 & 0.506$\pm$0.088 & 0.507$\pm$0.089 & 0.507$\pm$0.089 \\
&& Aff-F1 & 0.748$\pm$0.075 & 0.749$\pm$0.075 & 0.747$\pm$0.071 & 0.748$\pm$0.072 & 0.695$\pm$0.039 & 0.698$\pm$0.043 & 0.694$\pm$0.037 & 0.694$\pm$0.038 & 0.691$\pm$0.138 & 0.693$\pm$0.139 & 0.493$\pm$0.139 & 0.694$\pm$0.140 \\
\hhline{~|--------------}
& \multirow{4}{*}{\rotatebox[origin=c]{90}{\textbf{AnomTran}~\cite{xu2022anomaly}}} &
F1 & 0.295 & 0.307 & 0.304 & 0.306 & 0.345 & 0.347 & 0.343 & 0.346 & 0.402 & 0.402 & 0.409 & 0.398 \\
&& AUPR & 0.270$\pm$0.235 & 0.280$\pm$0.233 & 0.272$\pm$0.236 & 0.280$\pm$0.233 & 0.240$\pm$0.240 & 0.243$\pm$0.239 & 0.224$\pm$0.232 & 0.236$\pm$0.241 & 0.263$\pm$0.318 & 0.252$\pm$0.305 & 0.267$\pm$0.317 & 0.252$\pm$0.305 \\
&& AUROC & 0.761$\pm$0.142 & 0.760$\pm$0.142 & 0.768$\pm$0.142 & 0.759$\pm$0.140 & 0.633$\pm$0.142 & 0.637$\pm$0.147 & 0.631$\pm$0.140 & 0.641$\pm$0.148 & 0.654$\pm$0.195 & 0.647$\pm$0.190 & 0.652$\pm$0.195 & 0.648$\pm$0.191 \\
&& Aff-F1 & 0.659$\pm$0.241 & 0.660$\pm$0.242 & 0.665$\pm$0.240 & 0.663$\pm$0.239 & \underline{0.770$\pm$0.106} & \underline{0.768$\pm$0.103} & \underline{0.772$\pm$0.103} & \underline{0.772$\pm$0.104} & \underline{0.766$\pm$0.164}  & 0.764$\pm$0.162 & 0.765$\pm$0.163 & 0.765$\pm$0.163 \\
\hhline{~|--------------}
& \multirow{4}{*}{\rotatebox[origin=c]{90}{\textbf{DCdetector}~\cite{yang2023dcdetector}}} &
F1 & 0.079 & 0.085 & 0.085 & 0.083 & 0.208 & 0.215 & 0.203 & 0.201 & 0.280 & 0.283 & 0.287 & 0.278 \\
&& AUPR & 0.041$\pm$0.036 & 0.044$\pm$0.037 & 0.044$\pm$0.037 & 0.044$\pm$0.037 & 0.124$\pm$0.139 & 0.125$\pm$0.138 & 0.122$\pm$0.140 & 0.122$\pm$0.140 & 0.128$\pm$0.158 & 0.129$\pm$0.159 & 0.132$\pm$0.159 & 0.127$\pm$0.157 \\
&& AUROC & 0.496$\pm$0.008 & 0.495$\pm$0.008 & 0.495$\pm$0.008 & 0.495$\pm$0.008 & 0.501$\pm$0.011 & 0.501$\pm$0.011 & 0.501$\pm$0.011 & 0.501$\pm$0.011 & 0.516$\pm$0.082 & 0.513$\pm$0.079 & 0.516$\pm$0.082 & 0.517$\pm$0.082 \\
&& Aff-F1 & 0.674$\pm$0.008 & 0.675$\pm$0.009 & 0.675$\pm$0.009 & 0.675$\pm$0.010 & 0.706$\pm$0.035 & 0.707$\pm$0.035 & 0.707$\pm$0.035 & 0.708$\pm$0.037 & 0.678$\pm$0.075 & 0.680$\pm$0.076 & 0.670$\pm$0.074 & 0.680$\pm$0.075 \\
\hhline{~|--------------}
&\multirow{4}{*}{\rotatebox[origin=c]{90}{\textbf{DACAD}}} &
F1 & \textbf{0.600} & \textbf{0.633} & \textbf{0.598} & \textbf{0.572} & \textbf{0.595} & \textbf{0.528} & \textbf{0.522} & \textbf{0.528} & \textbf{0.532} & \textbf{0.464} & \textbf{0.651} & \textbf{0.463} \\
&& AUPR & \textbf{0.528}$\pm$\textbf{0.245} & \textbf{0.605}$\pm$\textbf{0.196} & \textbf{0.565}$\pm$\textbf{0.211} & \textbf{0.535}$\pm$\textbf{0.219} & \textbf{0.554}$\pm$\textbf{0.268} & \textbf{0.448}$\pm$\textbf{0.280} & \textbf{0.475}$\pm$\textbf{0.304} & \textbf{0.514}$\pm$\textbf{0.273} & \textbf{0.483}$\pm$\textbf{0.288} & \textbf{0.447}$\pm$\textbf{0.270} & \textbf{0.550}$\pm$\textbf{0.270} & \textbf{0.388}$\pm$\textbf{0.270} \\
&& AUROC & \textbf{0.856}$\pm$\textbf{0.094} & \textbf{0.858}$\pm$\textbf{0.083} & \textbf{0.822}$\pm$\textbf{0.093} & \textbf{0.813}$\pm$\textbf{0.098} & \textbf{0.787}$\pm$\textbf{0.268} & \textbf{0.748}$\pm$\textbf{0.142} & \textbf{0.719}$\pm$\textbf{0.191} & \textbf{0.769}$\pm$\textbf{0.151} & \textbf{0.760}$\pm$\textbf{0.151} & \textbf{0.687}$\pm$\textbf{0.202} & \textbf{0.790}$\pm$\textbf{0.105} & \textbf{0.619}$\pm$\textbf{0.234} \\
&& Aff-F1 & \underline{0.834$\pm$0.121} & \textbf{0.845$\pm$0.122} & \textbf{0.848$\pm$0.121} & \textbf{0.845$\pm$0.116}  & \textbf{0.814$\pm$0.123} & \textbf{0.785$\pm$0.119} & \textbf{0.804$\pm$0.125} & \textbf{0.810$\pm$0.125} & \textbf{0.785$\pm$0.133} & \textbf{0.798$\pm$0.143} & \textbf{0.799$\pm$0.149} & \textbf{0.798$\pm$0.149} \\
\hline
\end{tabular}%
}
\vspace{-10pt}
\end{table*}
\subsection{Baselines Comparison} \label{sec:compare}
Table \ref{tab:base} provides a comprehensive comparison of the performance of the different models on MTS benchmark datasets. The performance measurements include the F1 score (best F1 score), AUPR (Area Under the Precision-Recall Curve), and AUROC (Area Under the Receiver Operating Characteristic Curve). Despite Point Adjustment (PA) popularity in recent years, we do not use PA when calculating these metrics due to Siwon Kim's findings ~\cite{kim2022towards} that its application leads to an overestimation of a TSAD model's capability and can bias results in favor of methods that produce extreme anomaly scores. Instead, we use conventional performance metrics for anomaly detection.

Benchmark datasets like SMD contain multiple time series that cannot be merged due to missing timestamps, making them unsuitable for averaging their F1 scores. The F1 score is a non-additive metric combining precision and recall. To address this, we compute individual confusion matrices for each time series. These matrices are then aggregated into a collective confusion matrix for the entire dataset. From this aggregated matrix, we calculate the overall precision, recall, and F1 score, ensuring an accurate and undistorted representation of the dataset's F1 score. Additionally, we report the Affiliation F1 (Aff-F1)~\cite{huet2022local}, calculated from Aff-Precision (Aff-Pre) and Aff-Recall (Aff-Rec), to provide a comprehensive evaluation focused on early detection. For datasets with multiple time series, we present the mean and standard deviation of Aff-F1, AUPR, and AUROC for each series.

It is evident from Table \ref{tab:base} that DACAD is the best performer, consistently achieving the best results across all scenarios and metrics, as highlighted in bold. This suggests its robustness and adaptability to handling normal and anomalous representations of time series for anomaly detection. 
VRADA and CLUDA models often rank as the second best, with their results underlined in several instances. Other models like OmniAnomaly, THOC, TS2Vec, AnomalyTransformer, and DCdetector demonstrate more uniform performance across various metrics but generally do not reach the top performance levels seen in DACAD, VRADA, or CLUDA. Their consistent but lower performance could make them suitable for applications where top-tier accuracy is less critical.

The addition of the Affiliation F1 (Aff-F1) metric further reinforces the strength of DACAD. A high value of the affiliation metric indicates a strong overlap or alignment between the detected anomalies and the true anomalies in a time series. DACAD's superior Aff-F1 scores across all datasets underline its efficacy in accurately identifying anomalies, highlighting its potential as a solution in time series anomaly detection.

\begin{table*}[t]
\caption{Macro F1/AUROC results on SMD and Boiler dataset. The most optimal evaluation results are displayed in bold, while the second-best ones are indicated by \underline{underline}.}
\label{tab:context}
\centering
\footnotesize
\resizebox{0.8\linewidth}{!}{%
\renewcommand{\arraystretch}{1.2}
    \begin{tabular}{l|>{\centering}p{1.5cm}|>{\centering}p{1.5cm}|>{\centering}p{1.5cm}|>{\centering}p{1.5cm}|>{\centering}p{1.5cm}|>{\centering}p{1.5cm}|>{\centering}p{1.5cm}|>{\centering}p{1.5cm}|>{\centering\arraybackslash}p{1.5cm}}
    \hline
    \multirow{2}{*}{\rotatebox[origin=c]{90}{\textbf{Data}}}& & \multicolumn{8}{c}{\textbf{Models (Macro F1/AUROC)}} \\
    \hhline{~~--------}
    \textbf{} & \textbf{src $\mapsto$ trg} & \textbf{AE-MLP} & \textbf{AE-LSTM} & \textbf{RDC} & \textbf{VRADA} & \textbf{SASA} & \textbf{CLUDA} & \textbf{ContexTDA} & \textbf{DACAD} \\ 
    \hline
    \multirow{7}{*}{\rotatebox[origin=c]{90}{\textbf{SMD}}}
     & 1-1 $\mapsto$ 1-2 & 0.72 / 0.83 & 0.74 / 0.90 & 0.74 / 0.89 & 0.74 / 0.91 & 0.59 / 0.63 & \underline{0.75} / \underline{0.84} & \underline{0.75} / \textbf{0.91} & \textbf{0.80} / \underline{0.84} \\
     & 1-1 $\mapsto$ 1-3 & 0.57 / 0.70 & 0.49 / 0.41 & 0.57 / 0.72 & 0.49 / 0.41 & 0.61 / \textbf{0.90} & \underline{0.67} / 0.77 & 0.57 / 0.75 & \textbf{0.71} / \underline{0.84} \\
     & 1-1 $\mapsto$ 1-4 & 0.55 / 0.74 & 0.54 / 0.41 & 0.54 / 0.75 & 0.54 / 0.41 & 0.55 / 0.75 & 0.56 / 0.74 & \underline{0.59} / \underline{0.76} & \textbf{0.75} / \textbf{0.83} \\
     & 1-1 $\mapsto$ 1-5 & 0.54 / 0.79 & 0.55 / 0.84 & 0.56 / 0.85 & 0.55 / 0.79 & 0.65 / \underline{0.87} & \underline{0.68} / \underline{0.87} & 0.66 / \underline{0.87} & \textbf{0.82} / \textbf{0.96} \\
     & 1-1 $\mapsto$ 1-6 & 0.71 / 0.88 & 0.71 / 0.91 & 0.71 / 0.87 & 0.71 / \underline{0.91} & 0.44 / 0.84 & 0.71 / 0.77 & \underline{0.73} / 0.84 & \textbf{0.89} / \textbf{0.92} \\
     & 1-1 $\mapsto$ 1-7 & 0.48 / 0.55 & 0.48 / 0.50 & 0.49 / 0.54 & 0.48 / 0.50 & 0.31 / 0.57 & \underline{0.50} / \underline{0.62} & \underline{0.51} / 0.53 & \textbf{0.88} / \textbf{0.91} \\
     & 1-1 $\mapsto$ 1-8 & 0.55 / 0.57 & 0.53 / \underline{0.70} & 0.55 / 0.58 & 0.54 / 0.56 & 0.52 / 0.56 & 0.56 / 0.61 & \underline{0.58} / 0.58 & \textbf{0.79} / \textbf{0.72} \\
    \hline
    \multirow{6}{*}{\rotatebox[origin=c]{90}{\textbf{Boiler}}}
     & 1 $\mapsto$ 2 & 0.44 / 0.64 & 0.43 / 0.48 & 0.43 / 0.54 & 0.43 / 0.48 & \underline{0.53} / \textbf{0.88} & 0.49 / 0.52 & 0.50 / 0.59 & \textbf{0.61} / \underline{0.66} \\
     & 1 $\mapsto$ 3 & 0.40 / 0.28 & 0.40 / 0.11 & 0.43 / 0.49 & 0.42 / 0.18 & 0.41 / 0.48 & 0.47 / \underline{0.79} & \underline{0.50} / 0.67 & \textbf{0.74} / \textbf{0.81} \\
     & 2 $\mapsto$ 1 & 0.39 / 0.18 & 0.40 / 0.21 & 0.40 / 0.36 & 0.40 / 0.15 & \underline{0.53} / \textbf{0.90} & 0.47 / 0.49 & 0.51 / 0.66 & \textbf{0.58} / \underline{0.65} \\
     & 2 $\mapsto$ 3 & 0.40 / 0.38 & 0.40 / 0.20 & 0.45 / 0.39 & 0.42 / 0.21 & 0.41 / 0.49 & 0.48 / 0.50 & \underline{0.50} / \textbf{0.69} & \textbf{0.59} / \underline{0.63} \\
     & 3 $\mapsto$ 1 & 0.39 / 0.20 & 0.40 / 0.16 & 0.39 / 0.31 & 0.40 / 0.15 & 0.48 / 0.67 & 0.47 / \underline{0.77} & \underline{0.51} / 0.67 & \textbf{0.72} / \textbf{0.84} \\
     & 3 $\mapsto$ 2 & 0.48 / 0.54 & 0.49 / 0.48 & 0.49 / 0.55 & 0.49 / 0.48 & 0.46 / 0.31 & 0.49 / \underline{0.57} & \underline{0.50} / \underline{0.57} & \textbf{0.58} / \textbf{0.68} \\
    \hline
    \end{tabular}%
}
    \vspace{-10pt}
\end{table*}

\subsection{UDA Comparison}
Table \ref{tab:context}, adapted from~\cite{lai2023context}, now includes results of CLUDA and our model in the last two columns. This table provides a comprehensive comparison of the performances of various models on the SMD and Boiler datasets, as measured by Macro F1 and AUROC scores.

To assess our model against ContextDA \cite{lai2023context} --- the only existing UDA for the TSAD model --- we use the same datasets and metrics reported in ContextDA's main paper, as its code is unavailable. On the SMD dataset, ContextDA achieves an average macro F1 of 0.63 and AUROC of 0.75, whereas our model achieves a higher average macro F1 of 0.81 and AUROC of 0.86. Similarly, on the Boiler dataset, ContextDA's average macro F1 is 0.50 and AUROC 0.65, compared to our model's superior performance with an average macro F1 of 0.63 and AUROC of 0.71.
In this comparative analysis, the DACAD model distinctly outperforms others, consistently achieving the highest scores in both Macro F1 and AUROC across the majority of test cases in both datasets. Its performance is not only superior but also remarkably stable across different scenarios within each dataset, showcasing its robustness and adaptability to diverse data challenges.

The ContexTDA model frequently ranks as the runner-up after DACAD, particularly in the SMD dataset, where it often secures the second-highest scores in either or both the Macro F1 and AUROC metrics.
Interestingly, certain models exhibit a degree of specialization. For example, SASA, which underperforms in the SMD dataset, demonstrates notably better results in specific scenarios within the Boiler dataset, particularly in terms of the Macro F1 score. This suggests that some models may be more suited to specific types of data or scenarios.

In conclusion, while the DACAD model emerges as the most effective across most scenarios, the varying performances of different models across scenarios and datasets highlight the importance of carefully evaluating each model's unique characteristics and capabilities. A thoughtful selection process is essential to ensuring the most suitable model is chosen for a given task. This subtle approach is crucial in leveraging each model's strengths and achieving optimal results.

\subsection{Scalability Test} \label{sec:scalability}
\begin{figure}[t!]
    \centering
    \includegraphics[width=0.8\linewidth]{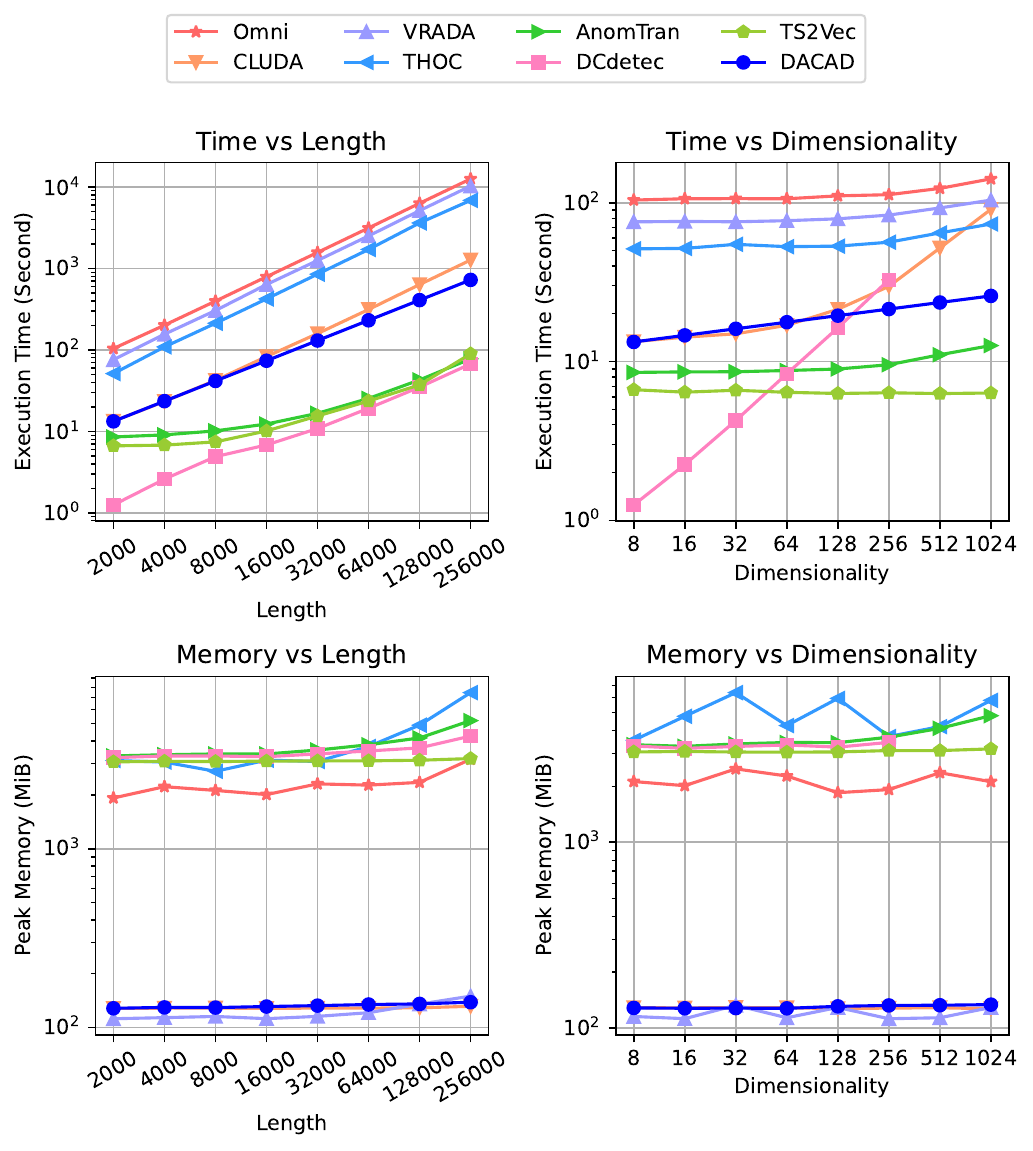}
    \caption{Comparison of execution time and peak memory usage for DACAD and other models. The top plots show execution time against time series length and dimensionality, while the bottom plots show memory usage.}
    \label{fig:time-memory}
    \vspace{-10pt}
\end{figure}

Scalability tests are crucial to evaluate a model's performance as data scales up in size and complexity. The plots in Fig.~\ref{fig:time-memory} provide a comprehensive comparison of DACAD's scalability with other models described in section~\ref{sec:compare}. The experiment measures execution time and peak memory usage in relation to the length of the time series and their dimensionality. For the dimensionality test, eight synthetic time series datasets with varying dimensions (ranging from 8 to 1024, doubling each step) and a fixed length of 2,000 are generated using the NeurIPS-TS~\cite{lai2021revisiting}. Another set of eight synthetic datasets with a fixed dimension of 8 varying lengths (ranging from 2,000 to 256,000) is created for the scale-up test concerning length. We report the execution time and memory usage of all models with the same parameter settings as in section~\ref{sec:compare}.

DACAD demonstrates excellent scalability in terms of time series length, outperforming other domain adaptation models such as VRADA and CLUDA. It maintains lower execution times, particularly evident with larger datasets, highlighting its efficiency. DACAD's scalability regarding time series dimensions is also remarkable. It remains stable as dimensionality increases, with execution times growing at a very small slope, ensuring it never exceeds reasonable memory limits. Compared to all other anomaly detection models, DACAD shows good scalability concerning time series length, consistently outperforming other methods and demonstrating its robustness in various performance metrics.

\subsection{Ablation Study} \label{sec:ablation}
Our ablation study focused on the following aspects: (1) \textbf{Effect of the loss components}, (2) \textbf{Effect of CEC}, and (3) \textbf{Effect of anomaly injection}.

\textbf{Effect of the loss components:}\label{subsec:els}
\begin{table}[t!]
\centering
\footnotesize
\renewcommand{\arraystretch}{1.2}
\caption{Effect of loss components on MSL dataset (source: F-5).}
\label{tab:loss}
    \begin{tabular}{c|ccc}
    \hline	
    Component & F1 & AUPR & AUROC\\
    \hline 
    DACAD & \textbf{0.595} & \textbf{0.554$\pm$0.268} & \textbf{0.787$\pm$0.268}  \\
    w/o $\mathcal{L}_{\text{SelfCont}}$ & 0.481 & 0.495$\pm$0.291 & 0.697$\pm$0.203 \\
    w/o $\mathcal{L}_{\text{SupCont}}$ & 0.463 & 0.427$\pm$0.274 & 0.639$\pm$0.198 \\
    w/o $\mathcal{L}_{\text{Disc}}$ & 0.471 & 0.484$\pm$0.315 & 0.699$\pm$0.229 \\
    w/o $\mathcal{L}_{\text{Cls}}$ & 0.299 & 0.170$\pm$0.127 & 0.503$\pm$0.018 \\
    \hline
    \end{tabular}
    \vspace{-10pt}
\end{table}
Table \ref{tab:loss} offers several key insights into the impact of different loss components on the MSL dataset using F-5 as a source. Removing the target self-supervised CL (w/o $\mathcal{L}_{\text{SelfCont}}$) leads to lower metrics (F1: 0.481, AUPR: 0.495, AUROC: 0.697). Moreover, excluding source-supervised CL (w/o $\mathcal{L}_{\text{SupCont}}$) reduces effectiveness (F1: 0.463, AUPR: 0.427, AUROC: 0.639), highlighting its role in capturing source-specific features, which are crucial for the model's overall accuracy. Similarly, omitting the discriminator component results in performance reduction (F1: 0.471, AUPR: 0.484, AUROC: 0.699). However, the most significant decline occurs without the classifier (F1: 0.299, AUPR: 0.170, AUROC: 0.503), underscoring its crucial role in effectively distinguishing between normal/anomaly classes. Overall, the best results (F1: 0.595, AUPR: 0.554, AUROC: 0.787) are achieved with all components, highlighting the effectiveness of an integrated approach.

Overall, each component within the model plays a crucial role in enhancing its overall performance. The highest performance across all metrics (F1: 0.595, AUPR: 0.554, AUROC: 0.787) is achieved when all components are included. 

\begin{figure}[t]
\centering
    \subfloat[\label{fig:tsne-src}]{
        \centering
        \includegraphics[width=0.2\textwidth]{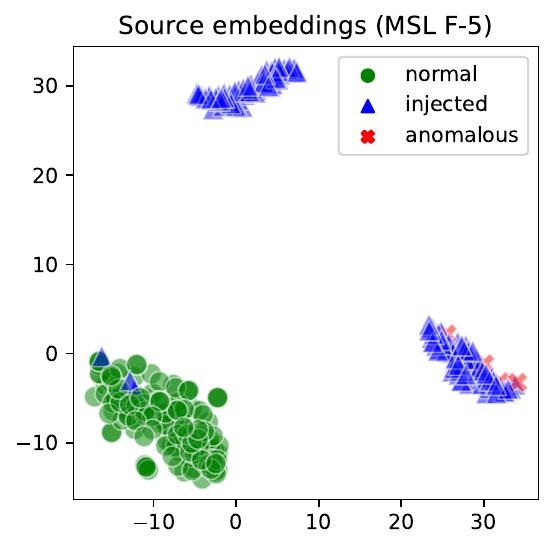}
        }
    \subfloat[\label{fig:tsne-trg}]{
        \centering
        \includegraphics[width=0.2\textwidth]{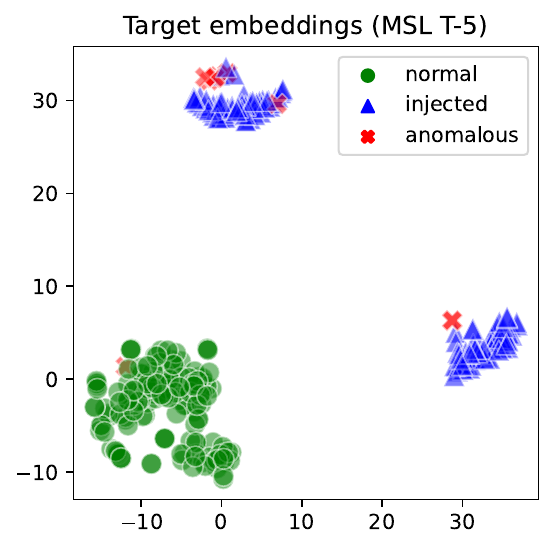}
        }
    \hfill
    \subfloat[\label{fig:tsne-src-uda}]{
        \centering
        \includegraphics[width=0.205\textwidth]{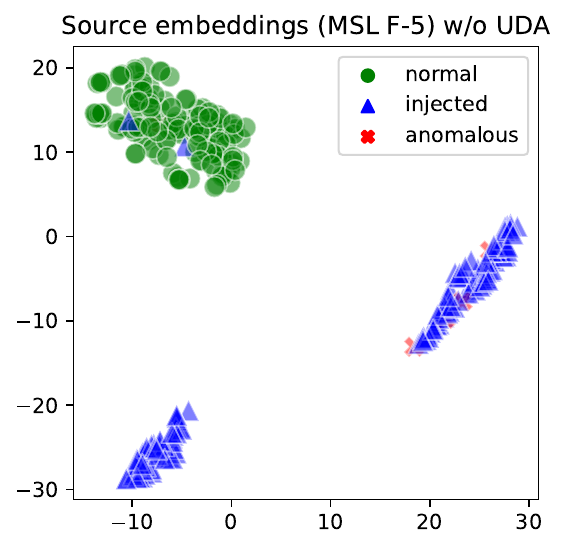}
        }
    \subfloat[\label{ffig:tsne-trg-uda}]{
        \centering
        \includegraphics[width=0.205\textwidth]{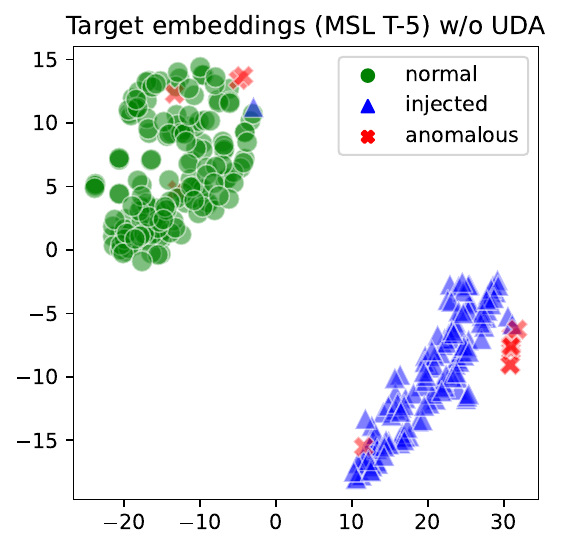}
        }
    \caption{Impact of UDA on DACAD feature representations. It contrasts the embeddings of (a) source entity F-5 with UDA, (b) target entity T-5 with UDA, (c) source entity F-5 without UDA, and (d) target entity T-5 without UDA. 
    }
    \label{fig:representations}
    \vspace{-10pt}
\end{figure}

To elucidate the effectiveness of UDA within DACAD, we examine the feature representations from the MSL dataset, as illustrated in Fig.~\ref{fig:representations}. It presents the t-SNE 2D embeddings of DACAD feature representations $\phi^{CL}(\phi^{R}(w))$ for the MSL dataset. Each point represents a time series window, which can be normal, anomalous, or anomaly-injected. These representations highlight the domain discrepancies between source and target entities and demonstrate how DACAD aligns the time series window features effectively. The comparison of feature representations with and without UDA reveals a significant domain shift when UDA is not employed between the source (entity F-5) and the target (entity T-5) within the MSL dataset.

\textbf{Effect of CEC:}\label{subsec:ecls}
Table \ref{tab:cls} compares the performance of our CEC classifier with two BCE and DeepSVDD - on the MSL dataset using F-5 as a source. 
Our proposed CEC shows superior performance compared to BCE and DeepSVDD across three metrics on the MSL dataset. With the highest F1 score, it demonstrates a better balance of precision and recall. Its leading performance in AUPR indicates greater effectiveness in identifying positive classes in imbalanced datasets. Additionally, CEC's higher AUROC suggests it is more capable of distinguishing between classes. 
\begin{table}[t!]
\centering
\footnotesize
\renewcommand{\arraystretch}{1.2}
\caption{Effect of CEC classifier on MSL dataset (source: F-5).}
\label{tab:cls}
    \begin{tabular}{c|ccc}
    \hline	
    Classifier & F1 & AUPR & AUROC\\
    \hline 
    Our proposed CEC & \textbf{0.595} & \textbf{0.554$\pm$0.268} & \textbf{0.787$\pm$0.268}  \\
    BCE-based & 0.467 & 0.504$\pm$0.304 & 0.670$\pm$0.241 \\
    DeepSVDD & 0.428 & 0.400$\pm$0.250 & 0.595$\pm$0.185 \\
    \hline
    \end{tabular}%
    \vspace{-10pt}
\end{table}

\begin{table}[t!]
\centering
\footnotesize
\renewcommand{\arraystretch}{1.2}
\caption{Effect of anomaly injection on MSL dataset (source: F-5).}
\label{tab:inj}
    \begin{tabular}{c|ccc}
    \hline	
    Approach & F1 & AUPR & AUROC\\
    \hline 
    with injection & \textbf{0.595} & \textbf{0.554$\pm$0.268} & \textbf{0.787$\pm$0.268}  \\
    w/o injection & 0.489 & 0.461$\pm$0.276 & 0.682$\pm$0.666 \\
    \hline
    \end{tabular}%
    \vspace{-10pt}
\end{table}

\textbf{Effect of anomaly injection:}\label{subsec:eai}
We study the impact of anomaly injection in Table \ref{tab:inj} on the MSL dataset when using F-5 as a source. It shows that anomaly injection significantly improves all metrics (F1: 0.595, AUPR: 0.554, AUROC: 0.787), enhancing the model's ability to differentiate between normal and anomalous patterns, thereby improving DACAD's overall accuracy. Without anomaly injection, there is a notable decline in performance, emphasizing its role in precision. The higher standard deviation in AUROC scores without injection suggests more variability and less stability in the model performance. The study underscores the vital role of anomaly injection in improving anomaly detection models. It reveals that incorporating anomaly injection not only boosts detection accuracy but also enhances the model's overall stability. 

\section{Conclusion}\label{sec:conclusion}
The DACAD model is an innovative approach to TSAD that is particularly effective in environments with limited labeled data.
By combining domain adaptation and contrastive learning, it applies labeled anomaly data from one domain to detect anomalies in another. Its anomaly injection mechanism, introducing a spectrum of synthetic anomalies, significantly bolsters the model's adaptability and robustness across various domains.
Our evaluations on diverse real-world datasets establish DACAD's superiority in handling domain shifts and outperforming existing models. Its capability to generalize and accurately detect anomalies, regardless of the scarcity of labeled data in the target domain, represents a significant contribution to TSAD.
In future work, we aim to refine the anomaly injection process further, enhancing the model's ability to simulate a broader range of anomalous patterns. Additionally, we plan to evolve the model to encompass univariate time series analysis, broadening its scope and utility. Furthermore, we intend to explore alternative feature extraction architectures beyond TCN to assess their impact on model performance.



\bibliographystyle{IEEEtran}
\bibliography{DACAD}

\begin{thebibliography}{10}
\providecommand{\url}[1]{#1}
\csname url@samestyle\endcsname
\providecommand{\newblock}{\relax}
\providecommand{\bibinfo}[2]{#2}
\providecommand{\BIBentrySTDinterwordspacing}{\spaceskip=0pt\relax}
\providecommand{\BIBentryALTinterwordstretchfactor}{4}
\providecommand{\BIBentryALTinterwordspacing}{\spaceskip=\fontdimen2\font plus
\BIBentryALTinterwordstretchfactor\fontdimen3\font minus \fontdimen4\font\relax}
\providecommand{\BIBforeignlanguage}[2]{{%
\expandafter\ifx\csname l@#1\endcsname\relax
\typeout{** WARNING: IEEEtran.bst: No hyphenation pattern has been}%
\typeout{** loaded for the language `#1'. Using the pattern for}%
\typeout{** the default language instead.}%
\else
\language=\csname l@#1\endcsname
\fi
#2}}
\providecommand{\BIBdecl}{\relax}
\BIBdecl

\bibitem{xie2022collaborative}
B.~Xie, S.~Li, F.~Lv, C.~H. Liu, G.~Wang, and D.~Wu, ``A collaborative alignment framework of transferable knowledge extraction for unsupervised domain adaptation,'' \emph{IEEE Transactions on Knowledge and Data Engineering}, vol.~35, no.~7, pp. 6518--6533, 2022.

\bibitem{liu2022deep}
X.~Liu, C.~Yoo, F.~Xing, H.~Oh, G.~El~Fakhri, J.-W. Kang, J.~Woo \emph{et~al.}, ``Deep unsupervised domain adaptation: A review of recent advances and perspectives,'' \emph{APSIPA Transactions on Signal and Information Processing}, vol.~11, 2022.

\bibitem{ismail2019deep}
H.~Ismail~Fawaz, G.~Forestier, J.~Weber, L.~Idoumghar, and P.-A. Muller, ``Deep learning for time series classification: a review,'' \emph{Data mining and knowledge discovery}, vol.~33, pp. 917--963, 2019.

\bibitem{wen2020time}
Q.~Wen, L.~Sun, F.~Yang, X.~Song, J.~Gao, X.~Wang, and H.~Xu, ``Time series data augmentation for deep learning: A survey,'' \emph{arXiv preprint arXiv:2002.12478}, 2020.

\bibitem{wilson2020survey}
G.~Wilson and D.~J. Cook, ``A survey of unsupervised deep domain adaptation,'' \emph{ACM Transactions on Intelligent Systems and Technology (TIST)}, vol.~11, pp. 1--46, 2020.

\bibitem{xu2024calibrated}
H.~Xu, Y.~Wang, S.~Jian, Q.~Liao, Y.~Wang, and G.~Pang, ``Calibrated one-class classification for unsupervised time series anomaly detection,'' \emph{IEEE Transactions on Knowledge and Data Engineering}, 2024.

\bibitem{purushotham2016variational}
S.~Purushotham, W.~Carvalho, T.~Nilanon, and Y.~Liu, ``Variational recurrent adversarial deep domain adaptation,'' in \emph{ICLR}, 2016.

\bibitem{cai2021time}
R.~Cai, J.~Chen, Z.~Li, W.~Chen, K.~Zhang, J.~Ye, Z.~Li, X.~Yang, and Z.~Zhang, ``Time series domain adaptation via sparse associative structure alignment,'' in \emph{AAAI}, vol.~35, 2021, pp. 6859--6867.

\bibitem{ozyurt2023contrastive}
Y.~Ozyurt, S.~Feuerriegel, and C.~Zhang, ``Contrastive learning for unsupervised domain adaptation of time series,'' \emph{ICLR}, 2023.

\bibitem{wilson2020multi}
G.~Wilson, J.~R. Doppa, and D.~J. Cook, ``Multi-source deep domain adaptation with weak supervision for time-series sensor data,'' in \emph{SIGKDD}, 2020, pp. 1768--1778.

\bibitem{zhang2022survey}
W.~Zhang, L.~Deng, L.~Zhang, and D.~Wu, ``A survey on negative transfer,'' \emph{IEEE/CAA Journal of Automatica Sinica}, vol.~10, no.~2, pp. 305--329, 2022.

\bibitem{zhou2024label}
Q.~Zhou, S.~He, H.~Liu, J.~Chen, and W.~Meng, ``Label-free multivariate time series anomaly detection,'' \emph{IEEE Transactions on Knowledge and Data Engineering}, 2024.

\bibitem{quintana2025bridging}
G.~I. Quintana, L.~Vancamberg, V.~Jugnon, A.~Desolneux, and M.~Mougeot, ``Bridging contrastive learning and domain adaptation: Theoretical perspective and practical application,'' \emph{arXiv preprint arXiv:2502.00052}, 2025.

\bibitem{lai2023context}
K.-H. Lai, L.~Wang, H.~Chen, K.~Zhou, F.~Wang, H.~Yang, and X.~Hu, ``Context-aware domain adaptation for time series anomaly detection,'' in \emph{Proceedings of the SDM'2023}.\hskip 1em plus 0.5em minus 0.4em\relax SIAM, 2023, pp. 676--684.

\bibitem{darban2023carla}
Z.~Z. Darban, G.~I. Webb, S.~Pan, C.~C. Aggarwal, and M.~Salehi, ``Carla: Self-supervised contrastive representation learning for time series anomaly detection,'' \emph{Pattern Recognition}, vol. 157, p. 110874, 2025.

\bibitem{ruff2018deep}
L.~Ruff, R.~Vandermeulen, N.~Goernitz, L.~Deecke, S.~A. Siddiqui, A.~Binder, E.~M{\"u}ller, and M.~Kloft, ``Deep one-class classification,'' in \emph{ICML}.\hskip 1em plus 0.5em minus 0.4em\relax PMLR, 2018, pp. 4393--4402.

\bibitem{li2023few}
H.~Li, W.~Zheng, F.~Tang, Y.~Zhu, and J.~Huang, ``Few-shot time-series anomaly detection with unsupervised domain adaptation,'' \emph{Information Sciences}, vol. 649, p. 119610, 2023.

\bibitem{long2018conditional}
M.~Long, Z.~Cao, J.~Wang, and M.~I. Jordan, ``Conditional adversarial domain adaptation,'' \emph{Advances in neural information processing systems}, vol.~31, 2018.

\bibitem{tzeng2017adversarial}
E.~Tzeng, J.~Hoffman, K.~Saenko, and T.~Darrell, ``Adversarial discriminative domain adaptation,'' in \emph{Proceedings of the IEEE conference on computer vision and pattern recognition}, 2017, pp. 7167--7176.

\bibitem{zhao2021unsupervised}
R.~Zhao, Y.~Xia, and Y.~Zhang, ``Unsupervised sleep staging system based on domain adaptation,'' \emph{Biomedical Signal Processing and Control}, vol.~69, p. 102937, 2021.

\bibitem{yang2021pipeline}
Y.~Yang, H.~Zhang, and Y.~Li, ``Long-distance pipeline safety early warning: a distributed optical fiber sensing semi-supervised learning method,'' \emph{IEEE sensors journal}, vol.~21, no.~17, pp. 19\,453--19\,461, 2021.

\bibitem{wang2021inter}
G.~Wang, M.~Chen, Z.~Ding, J.~Li, H.~Yang, and P.~Zhang, ``Inter-patient ecg arrhythmia heartbeat classification based on unsupervised domain adaptation,'' \emph{Neurocomputing}, vol. 454, pp. 339--349, 2021.

\bibitem{lu2021new}
N.~Lu, H.~Xiao, Y.~Sun, M.~Han, and Y.~Wang, ``A new method for intelligent fault diagnosis of machines based on unsupervised domain adaptation,'' \emph{Neurocomputing}, vol. 427, pp. 96--109, 2021.

\bibitem{ragab2020contrastive}
M.~Ragab, Z.~Chen, M.~Wu, C.~S. Foo, C.~K. Kwoh, R.~Yan, and X.~Li, ``Contrastive adversarial domain adaptation for machine remaining useful life prediction,'' \emph{IEEE Transactions on Industrial Informatics}, vol.~17, pp. 5239--5249, 2020.

\bibitem{hochreiter1997long}
S.~Hochreiter and J.~Schmidhuber, ``Long short-term memory,'' \emph{Neural computation}, vol.~9, pp. 1735--1780, 1997.

\bibitem{chung2015recurrent}
J.~Chung, K.~Kastner, L.~Dinh, K.~Goel, A.~C. Courville, and Y.~Bengio, ``A recurrent latent variable model for sequential data,'' \emph{Advances in neural information processing systems}, vol.~28, 2015.

\bibitem{zhou2023one}
T.~Zhou, P.~Niu, L.~Sun, R.~Jin \emph{et~al.}, ``One fits all: Power general time series analysis by pretrained lm,'' \emph{Advances in neural information processing systems}, vol.~36, pp. 43\,322--43\,355, 2023.

\bibitem{liu2024large}
J.~Liu, C.~Zhang, J.~Qian, M.~Ma, S.~Qin, C.~Bansal, Q.~Lin, S.~Rajmohan, and D.~Zhang, ``Large language models can deliver accurate and interpretable time series anomaly detection,'' \emph{arXiv preprint arXiv:2405.15370}, 2024.

\bibitem{zhang2024large}
X.~Zhang, R.~R. Chowdhury, R.~K. Gupta, and J.~Shang, ``Large language models for time series: A survey,'' \emph{arXiv preprint arXiv:2402.01801}, 2024.

\bibitem{jin2023large}
M.~Jin, Q.~Wen, Y.~Liang, C.~Zhang, S.~Xue, X.~Wang, J.~Zhang, Y.~Wang, H.~Chen, X.~Li \emph{et~al.}, ``Large models for time series and spatio-temporal data: A survey and outlook,'' \emph{arXiv preprint arXiv:2310.10196}, 2023.

\bibitem{yang2024survey}
Y.~Yang, M.~Jin, H.~Wen, C.~Zhang, Y.~Liang, L.~Ma, Y.~Wang, C.~Liu, B.~Yang, Z.~Xu \emph{et~al.}, ``A survey on diffusion models for time series and spatio-temporal data,'' \emph{arXiv preprint arXiv:2404.18886}, 2024.

\bibitem{wang2024drift}
C.~Wang, Z.~Zhuang, Q.~Qi, J.~Wang, X.~Wang, H.~Sun, and J.~Liao, ``Drift doesn't matter: Dynamic decomposition with diffusion reconstruction for unstable multivariate time series anomaly detection,'' \emph{Advances in Neural Information Processing Systems}, vol.~36, 2024.

\bibitem{zhao2024diffusion}
P.~Zhao, X.~Wang, Y.~Zhang, Y.~Li, H.~Wang, and Y.~Yang, ``Diffusion-uda: Diffusion-based unsupervised domain adaptation for submersible fault diagnosis,'' \emph{Electronics Letters}, vol.~60, no.~3, p. e13122, 2024.

\bibitem{schmidl2022anomaly}
S.~Schmidl, P.~Wenig, and T.~Papenbrock, ``Anomaly detection in time series: a comprehensive evaluation,'' \emph{Proceedings of the VLDB}, vol.~15, pp. 1779--1797, 2022.

\bibitem{audibert2022deep}
J.~Audibert, P.~Michiardi, F.~Guyard, S.~Marti, and M.~A. Zuluaga, ``Do deep neural networks contribute to multivariate time series anomaly detection?'' \emph{Pattern Recognition}, vol. 132, p. 108945, 2022.

\bibitem{xu2023deep}
H.~Xu, G.~Pang, Y.~Wang, and Y.~Wang, ``Deep isolation forest for anomaly detection,'' \emph{IEEE Transactions on Knowledge and Data Engineering}, vol.~35, no.~12, pp. 12\,591--12\,604, 2023.

\bibitem{darban2022deep}
Z.~Zamanzadeh~Darban, G.~I. Webb, S.~Pan, C.~Aggarwal, and M.~Salehi, ``Deep learning for time series anomaly detection: A survey,'' \emph{ACM Computing Surveys}, vol.~57, no.~1, pp. 1--42, 2024.

\bibitem{yue2022ts2vec}
Z.~Yue, Y.~Wang, J.~Duan, T.~Yang, C.~Huang, Y.~Tong, and B.~Xu, ``Ts2vec: Towards universal representation of time series,'' in \emph{AAAI}, vol.~36, 2022, pp. 8980--8987.

\bibitem{hundman2018detecting}
K.~Hundman, V.~Constantinou, C.~Laporte, I.~Colwell, and T.~Soderstrom, ``Detecting spacecraft anomalies using lstms and nonparametric dynamic thresholding,'' in \emph{SIGKDD}, 2018, pp. 387--395.

\bibitem{audibert2020usad}
J.~Audibert, P.~Michiardi, F.~Guyard, S.~Marti, and M.~A. Zuluaga, ``Usad: Unsupervised anomaly detection on multivariate time series,'' in \emph{SIGKDD}, 2020, pp. 3395--3404.

\bibitem{xu2022anomaly}
\BIBentryALTinterwordspacing
J.~Xu, H.~Wu, J.~Wang, and M.~Long, ``Anomaly transformer: Time series anomaly detection with association discrepancy,'' in \emph{ICLR}, 2022. [Online]. Available: \url{https://openreview.net/forum?id=LzQQ89U1qm_}
\BIBentrySTDinterwordspacing

\bibitem{niu2020lstm}
Z.~Niu, K.~Yu, and X.~Wu, ``Lstm-based vae-gan for time-series anomaly detection,'' \emph{Sensors}, vol.~20, p. 3738, 2020.

\bibitem{zhan2024hfn}
J.~Zhan, C.~Wu, C.~Yang, Q.~Miao, and X.~Ma, ``Hfn: Heterogeneous feature network for multivariate time series anomaly detection,'' \emph{Information Sciences}, vol. 670, p. 120626, 2024.

\bibitem{zhou2023semi}
F.~Zhou, G.~Wang, K.~Zhang, S.~Liu, and T.~Zhong, ``Semi-supervised anomaly detection via neural process,'' \emph{IEEE Transactions on Knowledge and Data Engineering}, vol.~35, no.~10, pp. 10\,423--10\,435, 2023.

\bibitem{su2019robust}
Y.~Su, Y.~Zhao, C.~Niu, R.~Liu, W.~Sun, and D.~Pei, ``Robust anomaly detection for multivariate time series through stochastic recurrent neural network,'' in \emph{SIGKDD}, 2019, pp. 2828--2837.

\bibitem{deng2021graph}
H.~Deng, Y.~Sun, M.~Qiu, C.~Zhou, and Z.~Chen, ``Graph neural network-based anomaly detection in multivariate time series data,'' in \emph{COMPSAC'2021}.\hskip 1em plus 0.5em minus 0.4em\relax IEEE, 2021, pp. 1128--1133.

\bibitem{shen2020timeseries}
L.~Shen, Z.~Li, and J.~Kwok, ``Timeseries anomaly detection using temporal hierarchical one-class network,'' \emph{Advances in Neural Information Processing Systems}, vol.~33, pp. 13\,016--13\,026, 2020.

\bibitem{lei2019}
Q.~Lei, J.~Yi, R.~Vaculin, L.~Wu, and I.~S. Dhillon, ``Similarity preserving representation learning for time series clustering,'' in \emph{IJCAI}, vol.~19, 2019, pp. 2845--2851.

\bibitem{zerveas2021}
G.~Zerveas, S.~Jayaraman, D.~Patel, A.~Bhamidipaty, and C.~Eickhoff, ``A transformer-based framework for multivariate time series representation learning,'' in \emph{SIGKDD}.\hskip 1em plus 0.5em minus 0.4em\relax Association for Computing Machinery, 2021, pp. 2114--2124.

\bibitem{tonekaboni2021}
S.~Tonekaboni, D.~Eytan, and A.~Goldenberg, ``Unsupervised representation learning for time series with temporal neighborhood coding,'' in \emph{ICLR}, 2021.

\bibitem{zhou2022contrastive}
H.~Zhou, K.~Yu, X.~Zhang, G.~Wu, and A.~Yazidi, ``Contrastive autoencoder for anomaly detection in multivariate time series,'' \emph{Information Sciences}, vol. 610, pp. 266--280, 2022.

\bibitem{yang2023dcdetector}
Y.~Yang, C.~Zhang, T.~Zhou, Q.~Wen, and L.~Sun, ``Dcdetector: Dual attention contrastive representation learning for time series anomaly detection,'' in \emph{SIGKDD}, 2023.

\bibitem{zhong2024patchad}
Z.~Zhong, Z.~Yu, Y.~Yang, W.~Wang, and K.~Yang, ``Patchad: A lightweight patch-based mlp-mixer for time series anomaly detection,'' \emph{arXiv preprint arXiv:2401.09793}, 2024.

\bibitem{schroff2015facenet}
F.~Schroff, D.~Kalenichenko, and J.~Philbin, ``Facenet: A unified embedding for face recognition and clustering,'' in \emph{Proceedings of the IEEE conference on computer vision and pattern recognition}, 2015, pp. 815--823.

\bibitem{khosla2020supervised}
P.~Khosla, P.~Teterwak, C.~Wang, A.~Sarna, Y.~Tian, P.~Isola, A.~Maschinot, C.~Liu, and D.~Krishnan, ``Supervised contrastive learning,'' \emph{Advances in neural information processing systems}, vol.~33, pp. 18\,661--18\,673, 2020.

\bibitem{lea2016temporal}
C.~Lea, R.~Vidal, A.~Reiter, and G.~D. Hager, ``Temporal convolutional networks: A unified approach to action segmentation,'' in \emph{Computer Vision--ECCV 2016 Workshops: Amsterdam, The Netherlands, October 8-10 and 15-16, 2016, Proceedings, Part III 14}.\hskip 1em plus 0.5em minus 0.4em\relax Springer, 2016, pp. 47--54.

\bibitem{creswell2018generative}
A.~Creswell, T.~White, V.~Dumoulin, K.~Arulkumaran, B.~Sengupta, and A.~A. Bharath, ``Generative adversarial networks: An overview,'' \emph{IEEE signal processing magazine}, vol.~35, no.~1, pp. 53--65, 2018.

\bibitem{ganin2015unsupervised}
Y.~Ganin and V.~Lempitsky, ``Unsupervised domain adaptation by backpropagation,'' in \emph{ICML}.\hskip 1em plus 0.5em minus 0.4em\relax PMLR, 2015, pp. 1180--1189.

\bibitem{bottou2018optimization}
L.~Bottou, F.~E. Curtis, and J.~Nocedal, ``Optimization methods for large-scale machine learning,'' \emph{SIAM review}, vol.~60, no.~2, pp. 223--311, 2018.

\bibitem{ghadimi2013stochastic}
S.~Ghadimi and G.~Lan, ``Stochastic first-and zeroth-order methods for nonconvex stochastic programming,'' \emph{SIAM journal on optimization}, vol.~23, no.~4, pp. 2341--2368, 2013.

\bibitem{sakurada2014anomaly}
M.~Sakurada and T.~Yairi, ``Anomaly detection using autoencoders with nonlinear dimensionality reduction,'' in \emph{Proceedings of the MLSDA 2014 2nd Workshop on Machine Learning for Sensory Data Analysis}, 2014, pp. 4--11.

\bibitem{malhotra2016lstm}
P.~Malhotra, L.~Vig, G.~Shroff, and P.~Agarwal, ``Lstm-based encoder-decoder for multi-sensor anomaly detection,'' in \emph{ICML}, 2016, pp. 1724--1732.

\bibitem{tzeng2014deep}
E.~Tzeng, J.~Hoffman, N.~Zhang, K.~Saenko, and T.~Darrell, ``Deep domain confusion: Maximizing for domain invariance,'' \emph{arXiv preprint arXiv:1412.3474}, 2014.

\bibitem{kim2022towards}
S.~Kim, K.~Choi, H.-S. Choi, B.~Lee, and S.~Yoon, ``Towards a rigorous evaluation of time-series anomaly detection,'' in \emph{AAAI}, vol.~36, 2022, pp. 7194--7201.

\bibitem{huet2022local}
A.~Huet, J.~M. Navarro, and D.~Rossi, ``Local evaluation of time series anomaly detection algorithms,'' in \emph{Proceedings of the 28th ACM SIGKDD Conference on Knowledge Discovery and Data Mining}, 2022, pp. 635--645.

\bibitem{lai2021revisiting}
K.-H. Lai, D.~Zha, J.~Xu, Y.~Zhao, G.~Wang, and X.~Hu, ``Revisiting time series outlier detection: Definitions and benchmarks,'' in \emph{Thirty-fifth Conference on Neural Information Processing Systems Datasets and Benchmarks Track (Round 1)}, 2021.

\end{thebibliography}

\end{document}